\documentclass[11pt]{article}

\usepackage[preprint]{acl}

\usepackage{times}
\usepackage{latexsym}
\usepackage{colortbl}
\definecolor{featLang}{HTML}{C8E6C9}   % soft green
\definecolor{featModel}{HTML}{BBDEFB}  % soft blue
\usepackage{url}
\usepackage{enumitem}
\usepackage[T1]{fontenc}
\usepackage[utf8]{inputenc}

\usepackage{microtype}

\usepackage{inconsolata}

\usepackage{graphicx}

\usepackage{amsmath}

\usepackage{booktabs}
\usepackage{array}
\usepackage{tabularx}
\usepackage{multirow}
\usepackage{colortbl} % \rowcolor / \cellcolor for booktabs-friendly tinting

\usepackage{amssymb}
\usepackage{graphicx}
\usepackage{tikz}
\usepackage{pgfplots}
\pgfplotsset{compat=1.18}

\usepackage{fvextra}

\usepackage{longtable}

\definecolor{famQwen}{HTML}{cfe2f3}    % blue
\definecolor{famCohere}{HTML}{d9ead3}  % green
\definecolor{famGPT}{HTML}{fce5cd}     % orange
\definecolor{famSarvam}{HTML}{ead1dc}  % magenta
\definecolor{bucketNLU}{HTML}{fff2cc}    % soft yellow
\definecolor{bucketReason}{HTML}{d0e0e3} % soft teal
\definecolor{bucketNLG}{HTML}{d9d2e9}    % soft lavender
\definecolor{metricGood}{HTML}{e8f0e3} % very light green
\definecolor{metricBad}{HTML}{f4e3e3}  % very light red
\definecolor{metricBest}{HTML}{fff2b3} % soft yellow highlight for best in column
\newcommand{\famrow}[1]{\rowcolor{#1!60}}
\newcommand{\hup}{$\uparrow$}
\newcommand{\hdn}{$\downarrow$}

\title{DEPART: DEcomposing PARiTy across Multilingual LLMs
}

\author{
  \textbf{Manan Uppadhyay}\textsuperscript{$\heartsuit$}, 
  \textbf{Prashant Kodali}\textsuperscript{$\heartsuit$},
  \textbf{Pranjal Chitale}\textsuperscript{$\heartsuit$}, \\
  \textbf{Reshma Ramaprasad}\textsuperscript{$\heartsuit$}, 
  \textbf{Himanshu Beniwal}\textsuperscript{$\heartsuit$$\clubsuit$}\thanks{Work done during internship at MSR India.}, 
  \textbf{Sunayana Sitaram}\textsuperscript{$\heartsuit$} \\
  \textsuperscript{$\heartsuit$}Microsoft Research India, 
  \textsuperscript{$\clubsuit$}IIT Gandhinagar \\
  \texttt{\{t-muppadhyay, sunayana.sitaram\}@microsoft.com}
}
\begin{document}
\maketitle

\begin{abstract}
Multilingual Large Language Models (mLLMs) leaderboards report per-language accuracy but rarely explain why disparities emerge, leaving systemic biases unattributed and offering practitioners no actionable levers. We first establish that these gaps are systematic rather than artifacts of sampling noise via distribution-free Friedman and Kruskal--Wallis tests, then introduce a two-step Bayesian hierarchical framework that decomposes multilingual performance variance into interpretable components. First, isolating the variance attributable to language identity, we show that observable language features (script, family, typological distance) explain $R^2_{\text{ling}} = 79\%$ of this variance on understanding tasks and $92\%$ on reasoning, with a model's internal representational similarity to English emerging as the dominant predictor across both task buckets. Second, decomposing the full (model$\times$benchmark$\times$language) cube, we find that NLU and reasoning have fundamentally divergent variance profiles: model identity dominates understanding ($66.7\%$ of variance), whereas the benchmark$\times$model interaction dominates reasoning ($46.3\%$). Together these results recast multilingual evaluation from passive performance mapping into an explainable, diagnostic framework with concrete levers for targeting the root drivers of language disparity.
\end{abstract}
\section{Introduction}
Large Language Models (LLMs) are deployed worldwide, yet their performance varies sharply across the languages they serve, producing systematically worse outputs for speakers of low-resource languages \citep{Choudhury_Deshpande_2021, khanuja-etal-2023-evaluating}.
The field measures this gap using per-language tables, occasional Gini coefficients, and English-vs-X deltas, but does not yet explain it.
Rather than just measuring the size of the cross-language gap, this work addresses a deeper question: how much of the gap can be predicted before running a benchmark, versus how much remains as an irreducible per-checkpoint residual?
We answer this using observable language properties (script, family, typological distance, resource class) and cheap model-conditional probes (tokenizer fertility, internal representation similarity to English).
The answer determines where intervention is most consequential: feature-predictable disparity points to data, tokenization, and alignment choices made before final evaluation, while a large residual would point to checkpoint-level idiosyncrasies that only retraining can address.

Existing work addresses this question partially: fairness frameworks \citep{Choudhury_Deshpande_2021, khanuja-etal-2023-evaluating} reduce disparity to one-dimensional summary statistics (Gini, max--min); the closest methodological precedent \citep{hu-etal-2025-quantifying} fits an additive mixed-effects model that ranks languages by difficulty but cannot, by construction, identify model$\times$language interactions or admit linguistic features as covariates; intrinsic-similarity rankings \citep{Li_Shi_Liu_Yang_Payani_Liu_Du_2025} and proxy-LM forecasts \citep{anugraha-etal-2025-proxylm} contribute individual signals without a fairness-style decomposition (see Section~\ref{sec:related}).
What is missing is a single hierarchical decomposition that jointly attributes observed disparity to feature-predictable and residual components, with calibrated uncertainty on both.

\begin{figure*}
    \centering \includegraphics[width=\textwidth,keepaspectratio]{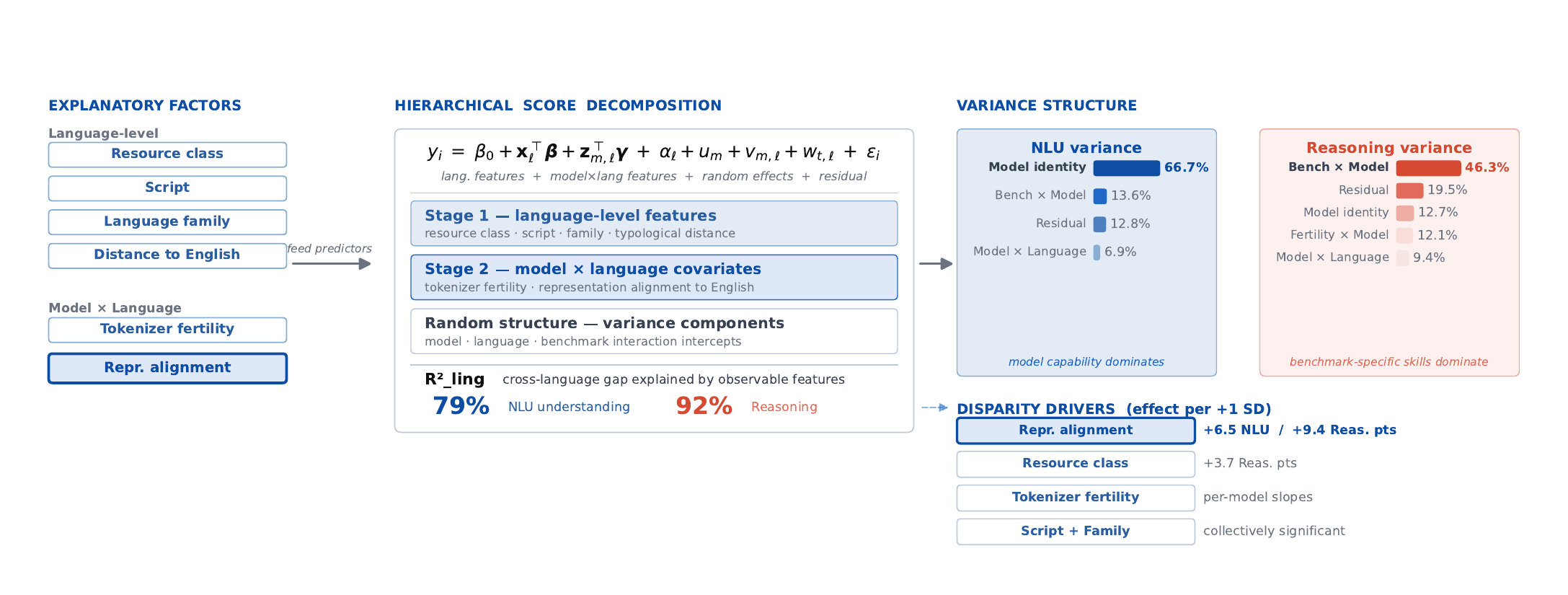}
    \caption{Structural factors dominate cross-lingual performance gaps in multilingual LLMs.}
    \label{fig:main}
\end{figure*}

We evaluate seven open-weight LLMs ($\sim$3.5B--122B parameters) on 15 multilingual benchmarks covering 63 languages, partitioned into multilingual-understanding (9 benchmarks, 63 languages) and reasoning (6 benchmarks, 62 languages) task buckets.
We fit a Bayesian hierarchical model that jointly estimates the contribution of the model, the benchmark, the language, their interactions, and observable language-level and model$\times$language features: resource class \citep{joshi-etal-2020-state}, lang2vec syntactic, phonological, and geographic distance to English \citep{littell-etal-2017-uriel}, script, family, tokenizer fertility, and the alignment between a model's representation of a language and its representation of English.
The analysis proceeds in two steps: distribution-free tests first verify that observed gaps are systematic, after which the hierarchical model decomposes them into feature-predictable and residual components.   

\noindent The following are the key contributions of this work.
\begin{itemize}[noitemsep, topsep=0pt]
    \item \textbf{Non-parametric validation of widespread language disparity.} Distribution-free tests verify that observed cross-lingual gaps are systematic rather than sampling artifacts: a Friedman test rejects equal language ranks across all 18 benchmarks, and a Dunn pairwise tests rejects performance equality across resource tiers, confirming that performance drops are structured by resource level.
    \item \textbf{A joint hierarchical decomposition with calibrated uncertainty.} We fit a Bayesian hierarchical model that estimates the cross-language gap (i.e.\ $\pm 8$--$9$ accuracy points) and the share explained by observable features, summarised by $R^2_{\text{ling}}$, the proportional reduction in $\sigma_\alpha^2$ once those features are added. Observable language features alone recover 0.79-0.92 $R^2$, showing that most cross-language variation is predictable in advance of evaluation. Across both task buckets, internal representation alignment to English is the single most dominant predictor (a one-SD increase is worth $+6.5$ accuracy points on understanding and $+9.4$ on reasoning), offering a benchmark-free proxy computable from model geometry alone.
    \item \textbf{Variance decomposition across the full design.} Model identity dominates understanding ($66.7\%$ of variance), whereas benchmark$\times$model interaction dominates reasoning ($46.3\%$, larger than the main model effect), so a single multilingual-reasoning aggregate is misleading and per-benchmark reporting is warranted.
\end{itemize}

% \paragraph{Non-parametric validation of cross-lingual disparity.}
% Distribution-free tests confirm that observed gaps are systematic: a Friedman test rejects equal language ranks across all benchmarks, and Dunn pairwise tests reject performance equality across resource tiers, establishing that drops are structured by resource level rather than sampling noise.

% \paragraph{Bayesian decomposition of cross-lingual variance.}
% A hierarchical model quantifies both the magnitude and predictability of language gaps. The cross-language posterior SD reveals consistent accuracy penalties across languages, and observable language features recover a large share of this variance, showing that most variation is predictable before evaluation. Internal representation alignment to English emerges as the dominant predictor, providing a benchmark-free proxy derivable from model geometry alone.

% \paragraph{Variance decomposition across the full design.}
% Model identity dominates understanding variance, but benchmark$\times$model interaction dominates reasoning---exceeding the main model effect---demonstrating that a single multilingual-reasoning aggregate is misleading and per-benchmark reporting is necessary.

\section{Related Works}
\label{sec:related}

\paragraph{Multilingual Evaluation Benchmarks.}
A growing body of multilingual evaluation benchmarks/suites for NLU tasks (XNLI, MEGA, MEGAVERSE, BUFFET) \citep{conneau-etal-2018-xnli,ahuja-etal-2023-mega,ahuja-etal-2024-megaverse,asai-etal-2024-buffet} have established
per-language reporting as the default,
with recent benchmarks like MMLU-ProX \citep{xuan-etal-2025-mmlu} and Global MMLU \citep{singh-etal-2025-global} pushing the language coverage across
knowledge / reasoning-intensive tasks, reading comprehension \citep{bandarkar-etal-2024-belebele}, and
pan-cultural settings like Include  \citep{romanou2025include}. These have also been complemented
by region-specific suites like MILU \citep{verma-etal-2025-milu},
Iroko Bench \citep{adelani-etal-2025-irokobench}, IndoMMLU \citep{koto-etal-2023-large}, SEA-HELM \citep{susanto-etal-2025-sea} for Indic, African, and South East Asian languages, respectively. 
While several of these works supplement per-language tables with \emph{bivariate} correlations against individual covariates like resource class \citep{ahuja-etal-2023-mega,asai-etal-2024-buffet}, pretraining data share, or tokenizer fertility \citep{ahuja-etal-2024-megaverse} - the dominant reporting format remains per-language accuracy together with an unweighted mean, with attribution typically performed one covariate at a time without controlling for confounders, modelling interactions, or quantifying uncertainty.

% Across this
% body of work, \textbf{the dominant reporting format remains per-language
% accuracy together with an unweighted mean, implying disparity is observed but rarely quantified, and never attributed.}

\paragraph{Disparity-aware Evaluation and its limits.}
\citet{Choudhury_Deshpande_2021} reframe multilingual model selection as
a social-choice problem and advocate Rawlsian max--min selection over
the implicit utilitarian means. \citet{khanuja-etal-2023-evaluating}
operationalize fairness as the Gini coefficient over per-language scores
within a Diversity--Equity--Inclusion framework, and
\citet{blasi-etal-2022-systematic} document the
systemic global inequalities in language-technology performance.
These works establish the normative case for disparity-aware
evaluation, but stop at one-dimensional summaries (Gini, max--min) and
do not run a confounder-controlled statistical model over the
(model$\,\times\,$benchmark$\,\times\,$language) cube. The closest
to our work is the work by \citet{hu-etal-2025-quantifying} who fit a
linear mixed-effects model over MEGA datasets and a panel of pre-2024
model variants and report a Performance Realisation Ratio per language. The additive specification
($s = \mu + \alpha_\ell + \beta_t + u_m + \varepsilon$) can rank
languages and tasks by difficulty but cannot, by construction, identify
model$\times$language interactions or admit linguistic features as
covariates.
\citet{Li_Shi_Liu_Yang_Payani_Liu_Du_2025} and
\citet{anugraha-etal-2025-proxylm} contribute, respectively, an
intrinsic similarity-to-English ranking and a proxy-LM forecast for
unseen (language, task) cells, but neither yields a fairness
decomposition with calibrated uncertainty.

\paragraph{Mechanisms of Multilingual Disparity.}
A separate body of literature has identified individual correlates of
cross-lingual performance: data scale and the resource level taxonomy
\citep{joshi-etal-2020-state,blevins-zettlemoyer-2022-language};
tokenizer fertility, which varies by orders of magnitude across
languages \citep{rust-etal-2021-good,ahia-etal-2023-languages} and translates directly into evaluation inequity via inflated token budgets and length-sensitive metrics \citep{petrov-etal-2023-language};
typological distance to English from URIEL/lang2vec
\citep{littell-etal-2017-uriel}; and the alignment of internal
representations with English, established for multilingual encoders
\citep{wu-dredze-2020-languages,conneau-etal-2020-unsupervised} and
recently extended to decoder LLMs
\citep{wendler-etal-2024-llamas,Li_Shi_Liu_Yang_Payani_Liu_Du_2025}.
Each correlate has been studied in isolation; none of these
threads enter all four families of predictors into a single
hierarchical model, reports their explanatory contributions on a common
scale, or quantifies how much of the cross-language gap is structural
and therefore predictable in advance of evaluation, which is exactly the gap
our framework addresses (Section~\ref{sec:disparity:method}).

\section{Experimental Setup} \label{sec:setup}

\paragraph{Models} \label{sec:models}
As shown in Table~\ref{tab:models}, we evaluate seven open-weight multilingual LLMs spanning  $\sim$3.5B - $\sim$122B parameters, covering three tokenizer families and four providers.

\begin{table}[t]
\centering
\small
\setlength{\tabcolsep}{4pt}
\renewcommand{\arraystretch}{1.1}
\resizebox{\columnwidth}{!}{%
\begin{tabular}{@{}llll@{}}
\toprule
\textbf{Model} & \textbf{Params} & \textbf{Arch.} & \textbf{Origin} \\
\midrule
\famrow{famQwen}   \textsc{Qwen3.5-4B}        & 4B         & Dense & Qwen \\
\famrow{famQwen}   \textsc{Qwen3.5-122B-A10B} & 122B (10B) & MoE   & Qwen \\
\famrow{famCohere} \textsc{aya-expanse-32b}   & 32B        & Dense & Cohere \\
\famrow{famCohere} \textsc{tiny-aya-global}   & 3.5B       & Dense & Cohere (distilled) \\
\famrow{famGPT}    \textsc{gpt-oss-20b}       & 20B        & MoE   & GPT-OSS \\
\famrow{famSarvam} \textsc{sarvam-30b}        & 30B        & MoE   & Sarvam (Indic) \\
\famrow{famSarvam} \textsc{sarvam-105b}       & 105B       & MoE   & Sarvam (Indic) \\
\bottomrule
\end{tabular}%
}
\caption{Models evaluated, grouped by family
(\colorbox{famQwen}{Qwen}, \colorbox{famCohere}{Cohere},
\colorbox{famGPT}{GPT-OSS}, \colorbox{famSarvam}{Sarvam}).}
\label{tab:models}
\end{table}

\begin{table}[t]
\centering
\small
\setlength{\tabcolsep}{4pt}
\renewcommand{\arraystretch}{1.15}
\newcommand{\bk}[1]{\textsc{#1}}

\begin{tabularx}{\columnwidth}{@{}
  >{\raggedright\arraybackslash}p{0.18\columnwidth}
  >{\raggedright\arraybackslash}X
  >{\raggedright\arraybackslash}p{0.20\columnwidth}@{}}
\toprule
\textbf{Bucket} & \textbf{Benchmarks} & \textbf{Score} \\
\midrule
\rowcolor{bucketNLU!55}
\textbf{NLU} \newline \textit{(MCQ, lik, knowledge centric)} &
\bk{global-mmlu}, \bk{mmlu-indic-roman}, \bk{mmlu-prox}, \bk{mmmlu},
\bk{okapi-mmlu}, \bk{milu}, \bk{include}, \bk{boolq-indic},
\bk{triviaqa-indic-mcq} &
\texttt{acc}~$\in[0,1]$ \\
\addlinespace
\rowcolor{bucketReason!55}
\textbf{Reasoning} \newline \textit{(gen.\ + MCQ \ + Infer)} &
\bk{gsm8k-indic}, \bk{mgsm} (CoT math, flexible exact-match);
\bk{belebele}, \bk{xcopa}, \bk{xstorycloze}, \bk{xwinograd}
(commonsense MCQ, likelihood) &
native~$[0,1]$ \\
\bottomrule
\end{tabularx}

\caption{Benchmark task buckets, grouped by skill
(\colorbox{bucketNLU}{NLU},
\colorbox{bucketReason}{Reasoning}). Each
$(\text{model},\text{benchmark},\text{language})$ cell is the
benchmark-level mean of subject-level scores. NLG benchmarks are
listed separately in Table~\ref{tab:buckets_nlg}.}
\label{tab:buckets}
\end{table}

\paragraph{Benchmarks and Task Buckets} \label{sec:benchmarks}
Cross-language scores are drawn from \textsc{lm-evaluation-harness}~\cite{eval-harness} runs against a consistent eval pipeline. Benchmarks are partitioned into three task buckets (Table~\ref{tab:buckets}) according to output format and scoring. The evaluation suite encompasses both $N$-way parallel datasets and structurally non-parallel benchmarks.

\paragraph{Features: Language \& Model$\times$Languages}
\label{sec:features}

Every $(\text{model},\text{language})$ cell carries two groups of predictors, listed in Appendix Table~\ref{tab:features}. The first group is purely language-level: \texttt{script} and \texttt{family} are categorical typological factors, \texttt{resource\_class} is the \citet{joshi-etal-2020-state} $1$--$5$ tier, and \texttt{syn\_dist\_en}, \texttt{phon\_dist\_en} are \texttt{lang2vec} distances to English. The second group is model-specific: \texttt{fertility} is tokens-per-word for the model's tokenizer, and \texttt{repr\_sim\_en} is the representational similarity (CKA) of the model's hidden-state geometry for the language to its geometry for English. All continuous features are standardised ($\mu=0$, $\sigma=1$) before entering the model.

\section{Characterizing Disparity} \label{sec:characterizing_disparity}

\begin{table}[t]
\centering
\small
\setlength{\tabcolsep}{4pt}
\renewcommand{\arraystretch}{1.12}

\resizebox{\columnwidth}{!}{%
\begin{tabular}{lrrl}
\toprule
\textbf{Benchmark} & $k$ (lang) & Kendall $W$ & sig. \\
\midrule
\rowcolor{bucketReason!55} \textsc{belebele}            & 62 & 0.696 & *** \\
\rowcolor{bucketNLU!55}    \textsc{boolq-indic}         & 11 & 0.810 & *** \\
\rowcolor{bucketNLU!55}    \textsc{global-mmlu}         & 15 & 0.728 & *** \\
\rowcolor{bucketReason!55} \textsc{gsm8k-indic}         & 11 & 0.808 & *** \\
\rowcolor{bucketNLU!55}    \textsc{include}             & 43 & 0.665 & *** \\
\rowcolor{bucketReason!55} \textsc{mgsm}                & 11 & 0.562 & *** \\
\rowcolor{bucketNLU!55}    \textsc{milu}                & 11 & 0.871 & *** \\
\rowcolor{bucketNLU!55}    \textsc{mmlu-indic-roman}    & 10 & 0.907 & *** \\
\rowcolor{bucketNLU!55}    \textsc{mmlu-prox}           & 29 & 0.633 & *** \\
\rowcolor{bucketNLU!55}    \textsc{mmmlu}               & 14 & 0.809 & *** \\
\rowcolor{bucketNLU!55}    \textsc{okapi-mmlu}          & 34 & 0.729 & *** \\
\rowcolor{bucketNLU!55}    \textsc{triviaqa-indic-mcq}  & 11 & 0.860 & *** \\
\rowcolor{bucketReason!55} \textsc{xcopa}               & 11 & 0.309 & **  \\
\rowcolor{bucketReason!55} \textsc{xstorycloze}         & 11 & 0.578 & *** \\
\rowcolor{bucketReason!55} \textsc{xwinograd}           &  6 & 0.778 & *** \\
\bottomrule
\end{tabular}%
}
\caption{Per-benchmark Friedman tests (languages as items, models as judges); larger Kendall's $W \in [0,1]$ indicates stronger inter-model agreement on language rankings. Rows tinted by bucket (\colorbox{bucketNLU}{NLU}, \colorbox{bucketReason}{Reasoning}); $**\,p<0.01$, $***\,p<0.001$. Full statistics in Table~\ref{tab:friedman_per_benchmark}.}
% \caption{Per-benchmark Friedman test results for the NLU and Reasoning buckets,
% with languages as items and models as judges. Kendall's $W \in [0,1]$ reports
% the effect-size analog, where larger values indicate stronger agreement among
% models in the induced language rankings. Rows are tinted by bucket
% (\colorbox{bucketNLU}{NLU}, \colorbox{bucketReason}{Reasoning}).
% Significance codes denote $p < 0.01$ ($**$) and $p < 0.001$ ($***$).
% A larger version of this table with the full test statistics is provided in
% Table~\ref{tab:friedman_per_benchmark}.}
\label{tab:friedman_per_benchmark_compact}
\end{table}

\paragraph{Overview of Disparity.}
In this work, we evaluate $N$ models on $B$ multilingual benchmarks, where each benchmark $b \in \{1, \dots, B\}$ covers a set of languages $\mathcal{L}_b$ with $|\mathcal{L}_b| = x_b$. Let $s_{n,b,\ell} \in \mathbb{R}$ denote the score of model $n$ on benchmark $b$ for language $\ell \in \mathcal{L}_b$. Beyond reporting raw performance, we are interested in characterizing the gap between the best- and worst-performing languages for a given model as well as more nuanced measures that summarise the full distribution of per-language scores $\{s_{n,b,\ell}\}_{\ell \in \mathcal{L}_b}$. Prior work has explored several factors contributing to such gaps, most notably the systematic under-representation of many of the world's languages in training data and benchmarks \citep{joshi-etal-2020-state, blasi-etal-2022-systematic}. While under-representation is a key driver of disparity, it is one of several confounding factors, alongside task difficulty, domain mismatch, and model capacity, that jointly shape observed cross-lingual performance gaps \citep{hu-etal-2023-systematic, ahuja-etal-2023-mega}. 

\paragraph{Existing Disparity Metrics.}
For a fixed model $n$ and benchmark $b$, let $\mathbf{s}=(s_1,\dots,s_K)$
be the vector of per-language scores with mean $\bar{s}$ and standard
deviation $\sigma_s$. Prior work summarises $\mathbf{s}$ with
scale-invariant dispersion indices: the coefficient of variation
$\mathrm{CV}=\sigma_s/\bar{s}$ \citep{hu-etal-2025-quantifying}, the
Gini coefficient $G\in[0,1]$ adapted from welfare economics
\citep{khanuja-etal-2023-evaluating, Choudhury_Deshpande_2021}, Sen
welfare $W=\bar{s}(1-G)$ \citep{10.2307/2296597} which combines level and
equity. Full definitions and per-model values
are given in Table~\ref{tab:model_bucket_disparity} (Appendix).

\paragraph{Shortcomings of these methods.}
Three limitations make these summaries inadequate as a stand-alone
account of disparity. First, CV and Gini are scale-invariant: small
numerical differences mask large absolute gaps---even the lowest-Gini 
model shows an average $\max$-$\min$ gap of roughly one-third the mean.
Second, most indices are insensitive to the overall performance level, so a
uniformly weak model and a uniformly strong one both register
``perfect parity''. Third, none identify \emph{which} factors---language,
task, model, or their interactions---drive the gaps. We therefore test
whether disparity is systematic (Section~\ref{sec:nonparam}) and attribute
it to interpretable factors via a Bayesian mixed-effects model
(Section~\ref{sec:whatexplains}).

\paragraph{Is the disparity real and systematic?}
\label{sec:nonparam}

The summary statistics above describe \emph{how much} scores spread
across languages, but not whether that spread reflects a stable,
model-independent ordering or merely sampling noise. We address this
with two distribution-free tests - Kendall's $W$ \cite{kendall1939problem} and Dunn's Post-Hoc \cite{8380ffda-f46c-385f-9bf1-f1fe8f8da043}-  that together motivate the
mixed-effects model of Section~\ref{sec:whatexplains}.

\paragraph{Rankings are consistent across models (Friedman / Kendall's $W$).}

\begin{table}[t]
\centering
\small
\setlength{\tabcolsep}{6pt}
\renewcommand{\arraystretch}{1.15}
\begin{tabular}{lccc}
\toprule
\textbf{Resource tier} &
\cellcolor{bucketNLU!55}\textbf{NLU}\,\hup &
\cellcolor{bucketReason!55}\textbf{Reasoning}\,\hup &
\textbf{Overall}\,\hup \\
\midrule
1-Scraping  & 0.432          & 0.382          & 0.421          \\
2-Hopefuls  & 0.458          & 0.497          & 0.472          \\
3-Rising    & 0.536          & 0.527          & 0.533          \\
4-Underdogs & 0.542          & 0.667          & 0.604          \\
5-Winners   & \textbf{0.680} & \textbf{0.714} & \textbf{0.696} \\
\bottomrule
\end{tabular}
\caption{Mean per-language score by language resource tier, pooled
across models, on the NLU and Reasoning buckets
(\colorbox{bucketNLU}{NLU},
\colorbox{bucketReason}{Reasoning}).}
\label{tab:tier_mean_score}
\end{table}

For each benchmark, we treat languages as $k$ items ranked by the
$m{=}7$ models and compute the Friedman statistic together with
Kendall's $W = \chi^2 / [m(k{-}1)] \in [0,1]$ as an effect size for
inter-model agreement (Tables ~\ref{tab:friedman_per_benchmark_compact} and ~\ref{tab:friedman_per_benchmark}), where $W$ is the Kendall's W value; $\chi^2$ is the Friedman test statistic value. Kendall’s $W$ coefficient assumes the value from 0 (indicating no relationship) to 1 (indicating a perfect relationship). Null hypothesis of equal language ranks is rejected in \textbf{18/18 benchmarks},
with $W$ ranging from $0.31$ (\texttt{xcopa}) to $0.91$
(\texttt{mmlu\_indic\_roman}) and a median of $\sim\!0.75$. This means the models show similar patterns of ranking languages from easiest to hardest, with the same languages tending to be on top and bottom across models.
% Agreement is highest on translation (\texttt{igb\_flores}, \texttt{igb\_xsum}, \texttt{in22\_conv\_16k}) and Indic MMLU variants ($W \geq 0.80$), where the easy-to-hard language ordering is essentially model-invariant.

\paragraph{Disparity tracks resource tiers (Dunn's post-hoc).}
We next ask whether the cross-language spread aligns with how
well-represented, each language is on the web. We group languages
into the five resource tiers of \citet{joshi-etal-2020-state}
(Class~1 \emph{Scraping} $\rightarrow$ Class~5 \emph{Winners}),
$z$-normalise scores within each benchmark cell to make them comparable, and run Dunn's
pairwise tests between tiers with Benjamini--Hochberg FDR
correction at $\alpha=0.05$ \citep{https://doi.org/10.1111/j.2517-6161.1995.tb02031.x} Pooled across models, mean
$z$-scores rise strictly from Class~1 to Class~5 and \emph{all}
$10$ pairwise tier contrasts are significant in both NLU and
reasoning. The same monotone staircase holds per model on NLU
(7--10/10 contrasts significant for every model) and on reasoning
for all but two models -- Sarvam-30b and GPT-OSS-20B -- whose
tier ordering flattens at the lower end, an expected outcome when a model
is near the floor across most languages. Full pairwise matrices are
in Appendix~\ref{app:dunn_full}. The raw (unnormalized) means in
Table~\ref{tab:tier_mean_score} make the effect concrete:The 
average per-language score nearly doubles from Class~1 to
Class~5 in both buckets.

Together, these tests establish that the cross-language gaps are
neither sampling noise nor an artifact of any single model: the
language ordering is stable across models and structured monotonically
by resource level. This rules out the most benign explanations for
disparity and motivates an explanatory model of \emph{what} drives
it, beyond any single dispersion summary.

\section{What Explains Disparity?}
\label{sec:whatexplains}
Sources of the disparity characterised in Section~\ref{sec:characterizing_disparity} are entangled: a low score on a given language may reflect the language itself (its script, its typological distance from training data), the model (its tokenizer, its representation geometry), the benchmark, or simple noise. Disentangling these requires a model that respects the nested structure of the data --- every score is jointly indexed by $(\text{model}, \text{benchmark}, \text{language})$ --- and that yields a quantitative attribution rather than per-language anecdote. We address this with a Bayesian hierarchical decomposition fit independently across two task buckets: NLU (multilingual MCQ knowledge benchmarks) and Reasoning (commonsense and math MCQ benchmarks).\footnote{A third bucket, NLG (open generation; chrF on translation and summarisation), is reported in Appendix~\ref{app:nlg} but excluded from the main analysis: its $12$-language Indic/Dravidian pool is too narrow to identify the categorical script and family effects, so the disparity-explanation metric is not interpretable.}

% \paragraph{Key Finding:} Static structural features of languages, i.e. resource class, typological distances, script, and family explain $\mathbf{79\%}$ of cross-language variance on NLU and $\mathbf{92\%}$ on Reasoning. The single most consistent driver, across task buckets, is the alignment of a model's internal representations with its
% representation of English: a one-standard-deviation increase in this similarity is worth $+6.5$ accuracy points on NLU and $+9.4$ on Reasoning, with credible intervals that exclude zero in every bucket we measured.

\subsection{Method in brief} \label{sec:disparity:method}

We model each observed score as the sum of a benchmark effect ($\tau_t$), a language effect ($\alpha_\ell$), a model effect ($u_m$), and Gaussian noise ($\varepsilon$):
\begin{equation}
\label{eq:base}
\text{score}_{m,t,\ell} \;=\; \mu + \tau_t + \alpha_\ell + u_m + \varepsilon_{m,t,\ell},
\end{equation}
with $\alpha_\ell \sim \mathcal{N}(0, \sigma_\alpha^2)$, $u_m \sim \mathcal{N}(0, \sigma_u^2)$, and $\varepsilon \sim \mathcal{N}(0, \sigma^2)$. 

The key quantity is $\sigma_\alpha$, the typical magnitude of a
language's deviation from the model-and-benchmark-conditional mean. It is the formal answer to ``how big is the cross-language gap?''

To attribute that gap, we extend (\ref{eq:base}) in two steps. \textbf{First}, we add only language-level covariates $\mathbf{x}_\ell$ (resource class~\citep{joshi-etal-2020-state}, syntactic, phonological, and geographic distance to English from \texttt{lang2vec}~\citep{littell-etal-2017-uriel}, plus $\mathrm{script}(\ell)$ and $\mathrm{family}(\ell)$ as categorical factors). This shrinks $\sigma_\alpha$ to a residual $\sigma_\alpha^{(\mathrm{lang})}$. \textbf{Second}, we add model--language covariates $\mathbf{z}_{m,\ell}$ (tokenizer fertility and representation similarity to English) together with the random interactions $(1 \mid \mathrm{model}{:}\mathrm{language})$, $(1 \mid \mathrm{bench}{:}\mathrm{language})$, $(1 \mid \mathrm{bench}{:}\mathrm{model})$, and a per-model random slope on fertility. We measure representational similarity to English by computing the Centered Kernel Alignment (CKA)~\citep{kornblith19a} distance between the hidden states of English and the target language at the model's middle layer following \citep{wendler-etal-2024-llamas, dumas-etal-2025-separating}, mean-pooled across tokens on the FLORES-200 dataset~\cite{costa2022no}. We measure the tokenizer fertility on FLORES-200 as well. This dataset was specifically selected because it offers expert-curated parallel sentences translated uniformly across 200 languages. We provide a detailed justification for the design choices in Appendix~\ref{app:design_choices}, and provide a glossary of all the terminology in Appendix Table~\ref{tab:var_glossary}. 

Putting both extensions together, each observation $i$ (a triple $(m(i), t(i), \ell(i))$ of model, benchmark, and language) is modelled as
\begin{equation}
\label{eq:full}
\begin{aligned}
y_i \;=\;& \underbrace{\beta_0 + \tau_{t(i)} + \mathbf{x}_i^{\!\top}\boldsymbol{\beta}}_{\text{fixed effects}} \;+\; \underbrace{\varepsilon_i}_{\text{residual}} \\
        +\;& \underbrace{\alpha_{\ell(i)} + u_{m(i)} + v_{m(i),\,\ell(i)} + w_{t(i),\,\ell(i)}}_{\text{random effects}},
\end{aligned}
\end{equation}
where $\tau_{t(i)}$ is the benchmark fixed effect carried over from~(\ref{eq:base}), $\mathbf{x}_i$ stacks the language- and model--language covariates introduced above ($\mathbf{x}_\ell$ and $\mathbf{z}_{m,\ell}$), $\alpha_\ell$, $u_m$, $v_{m,\ell}$, $w_{t,\ell}$ are zero-mean Gaussian random intercepts with their own variance components $\sigma_\alpha^2, \sigma_u^2, \sigma_v^2, \sigma_w^2$, and $\varepsilon_i \sim \mathcal{N}(0, \sigma^2)$. We place weakly informative priors on $\boldsymbol{\beta}$ and half-normal priors on the variance components, and perform full posterior inference, yielding credible intervals on every coefficient and variance component reported below.

Our disparity-explanation metric is the proportional reduction
in between-language variance contributed by the language covariates:
\begin{equation}
\label{eq:r2ling}
R^2_{\text{ling}} \;=\; 1 \;-\; \frac{\bigl[\sigma_\alpha^{(\mathrm{lang})}\bigr]^2}{\sigma_\alpha^2}.
\end{equation}
This is a standard proportional-reduction-in-variance measure in multilevel modelling; we use the $R^2_{\text{ling}}$ notation to emphasise that it quantifies linguistic-feature explanatory share. All fits use NUTS via \texttt{numpyro} with 5000 tuning iterations and 2000 post-warmup draws per chain across four chains.

\subsection{The language gap and what explains it}
\label{sec:disparity:r2ling}
Without any structured features, languages differ by
$\sigma_\alpha = 0.078$ on NLU and $0.088$ on Reasoning around the
model-and-benchmark-conditional mean indicating typical deviations of
roughly $\pm 8$--$9$ accuracy points (Table~\ref{tab:headline}).

Adding just the language covariates plus script and family
categoricals shrinks $\sigma_\alpha$ to $0.035$ on NLU and
$0.021$ on Reasoning, giving $R^2_{\text{ling}} = 0.79$
$[0.57, 0.93]$ on NLU and $R^2_{\text{ling}} = 0.92$
$[0.73, 0.99]$ on Reasoning ($90\%$ HDIs;
Table~\ref{tab:headline}). The bulk of the cross-language gap on
MCQ benchmarks is recoverable from features that can be computed without any inference on the model itself.

\begin{table}[t]
\centering
\small
\setlength{\tabcolsep}{6pt}
\renewcommand{\arraystretch}{1.2}
\begin{tabular}{lcc}
\toprule
\textbf{Bucket} & $\sigma_\alpha$\,\hdn & $R^2_{\text{ling}}$\,\hup \\
\midrule
\rowcolor{bucketNLU!55}    \textbf{NLU}       & $0.078$ $[0.064, 0.092]$ & $\mathbf{0.79}$ $[0.57, 0.93]$ \\
\rowcolor{bucketReason!55} \textbf{Reasoning} & $0.088$ $[0.072, 0.107]$ & $\mathbf{0.92}$ $[0.73, 0.99]$ \\
\bottomrule
\end{tabular}
\caption{Headline disparity-explanation results per bucket.
$\sigma_\alpha$ is the SD of the language random intercept;
$R^2_{\text{ling}}$ is its proportional reduction once linguistic
features are added. Brackets show $90\%$ HDIs.}
\label{tab:headline}
\end{table}

\paragraph{Which single predictor carries the most weight?}
$R^2_{\text{ling}}$ above is computed from the language-features
fit alone. The full model additionally enters two
\emph{model-conditional} covariates: tokenizer fertility and
representation similarity to English (\texttt{repr\_sim\_en}) which lets us rank all six standardised continuous predictors (four
language-only, two model-conditional) against each other on a
common scale. Only one is credibly positive in both buckets:
\texttt{repr\_sim\_en}, with $\beta = +0.065$ $[0.051, 0.079]$ on
NLU and $+0.094$ $[0.074, 0.116]$ on Reasoning ($90\%$ HDIs). A
one-SD increase in alignment is worth $6$--$9$ accuracy points.\footnote{The same effect holds on NLG (Appendix~\ref{app:nlg}, $\beta = +0.053$ $[0.025, 0.080]$), making this the single most stable cross-bucket finding, and a quantity that can be computed from a model alone without running any multilingual benchmark.}

Resource level (\texttt{resource\_class}) also has a credibly
positive effect on Reasoning ($\beta = +0.037$, $[0.002, 0.07]$).
The other three typological distances (syntactic, phonological,
geographic) and tokenizer fertility have credible intervals that
overlap zero on their own. This does \emph{not} mean these
features carry no signal but rather, they are collinear with the
categorical script and family terms and with the per-language
random intercept, which soak up the shared variance. The high
joint $R^2_{\text{ling}}$ confirms the linguistic block as a whole
is informative.

\subsection{What else drives score variation?}
\label{sec:disparity:variance}

$R^2_{\text{ling}}$ answers ``how much of the language gap do
language features explain?''. A second, related question is
``what \emph{other} sources of variation in the score data should we
care about, once we account for everything we can?''
Table~\ref{tab:variance} reports the full-model decomposition: per
component and per bucket, the share of total variance carried by
that component and its standard deviation in raw score units (a
component with $\sigma = 0.10$ means typical deviations of
$\pm 10$ accuracy points along that axis). Columns sum to $100\%$
of total score variance within each bucket.

\paragraph{Language Residual:}  
The language-intercept SD in this full-model decomposition
($\sigma_{\mathrm{language}} \approx 0.017$) is smaller than the
language-features residual reported in Table~\ref{tab:headline}
($0.021$--$0.035$). The two answer different questions:
$R^2_{\text{ling}}$ measures the shrinkage of $\sigma_\alpha$ when
language features are added to the baseline (Eq.~\ref{eq:r2ling});
the full model additionally includes $(\mathrm{model}{:}\mathrm{language})$
and $(\mathrm{bench}{:}\mathrm{language})$ random interactions,
which absorb cell-specific deviations that were previously charged
to language identity, shrinking what remains in the language
intercept further.

\begin{table}[t]
\centering
\small
\setlength{\tabcolsep}{6pt}
\renewcommand{\arraystretch}{1.15}
\begin{tabular}{l rr rr}
\toprule
& \multicolumn{2}{c}{\cellcolor{bucketNLU!55}\textbf{NLU}} 
& \multicolumn{2}{c}{\cellcolor{bucketReason!55}\textbf{Reasoning}} \\
\cmidrule(lr){2-3} \cmidrule(lr){4-5}
\textbf{Component} & \textbf{\%} & $\boldsymbol{\sigma}$ & \textbf{\%} & $\boldsymbol{\sigma}$ \\
\midrule
$\sigma_{\mathrm{model}}$
  & $\mathbf{66.7}$ & $0.149$ & $12.7$          & $0.057$ \\
$\sigma_{\mathrm{bench} \times \mathrm{model}}$
  & $13.6$ & $0.061$ & $\mathbf{46.3}$ & $0.119$ \\
$\sigma_{\mathrm{fert\,slope}\mid\mathrm{model}}$
  & $1.4$  & $0.015$ & $12.1$          & $0.058$ \\
\midrule
$\sigma_{\mathrm{model} \times \mathrm{language}}$
  & $6.9$  & $0.044$ & $9.4$           & $0.053$ \\
$\sigma_{\mathrm{bench} \times \mathrm{language}}$
  & $3.6$  & $0.031$ & $4.3$           & $0.036$ \\
$\sigma_{\mathrm{language}}$ (residual)
  & $1.5$  & $0.017$ & $1.4$           & $0.018$ \\
\midrule
$\sigma$ (per-row noise)
  & $6.3$  & $0.042$ & $13.8$          & $0.065$ \\
\bottomrule
\end{tabular}
\caption{Variance decomposition of the full model, per bucket.
``\%'' is each component's share of total variance (columns sum to
$100$ within bucket); $\sigma$ is the posterior-mean SD in raw score
units. Bold marks the largest component per column.}
\label{tab:variance}
\end{table}

\paragraph{NLU and Reasoning have qualitatively different variance
structures.} On NLU, model identity dominates: $66.7\%$ of variance,
$\sigma_{\mathrm{model}} = 0.149$, meaning models differ by about
$15$ accuracy points and a model strong on \texttt{mmlu} tends to
be strong on \texttt{include} too. On Reasoning the ordering
inverts. $\mathrm{bench} \times \mathrm{model}$ carries $46.3\%$
($\sigma = 0.119$) and is \emph{larger} than the main model effect
($\sigma_{\mathrm{model}} = 0.057$): the typical model-by-benchmark
interaction ($\sim 12$ accuracy points) exceeds the typical
model-by-grand-mean deviation ($\sim 6$ points). A model's strength
does not pool across reasoning benchmarks --- math, commonsense, and
coreference probe genuinely different capabilities. A single
``multilingual reasoning'' aggregate is misleading; per-benchmark
reporting is required.

\paragraph{Tokenizer fertility matters on Reasoning, but
heterogeneously across models.}
The pooled fixed effect of \texttt{fertility\_z} has an HDI that
just crosses zero (Appendix~\ref{app:coefs}), but the per-model
random slope carries $12.1\%$ of Reasoning variance
($\sigma = 0.058$). Per-model fertility coefficients differ by
$\sim \pm 6$ accuracy points around the pooled mean, so averaging
over models hides a real effect. The same component is small on
NLU ($1.4\%$), consistent with MCQ knowledge benchmarks being less
tokenization-sensitive than chain-of-thought reasoning.

\paragraph{Model x Language persists at
$7$--$9\%$ in both buckets.}
With $\sigma \approx 0.044$--$0.053$, some model--language pairings
have idiosyncratic gains or losses of $\sim 5$ accuracy points that
none of our measured features explain --- almost certainly traceable
to pretraining-data specifics we cannot observe.

\paragraph{Non-parametric cross-check.}
As a sanity check on the feature ranking above, a gradient-boosted
tree fit to the same per-cell frame reproduces the Stage-3 ordering
--- representation similarity to English dominant, fertility and
syntactic distance secondary, phonological distance and resource
class marginal --- and recovers held-out language accuracies within
$\sim 4$ (NLU) and $\sim 7$ (Reasoning) accuracy points
(Appendix~\ref{sec:tree_validation}).

\section{Discussion and Implications}

\paragraph{Most of the cross-language gap is structural.}
Conditional on the model and the benchmark, languages differ by about
$\pm 8$--$9$ accuracy points. Observable language features --
resource class, typological distances, script, family, and
representation alignment to English -- explain $79\%$ of that gap
on understanding and $92\%$ on reasoning. This has a practical
consequence: an absolute gain on a low-resource language that does
not also reduce $\sigma_\alpha$ after controlling for these
features is, with high probability, attributable to structural factors (Say tokenizer fertility and increased representational similarity) rather
than a real improvement in multilingual capability. Reporting
$\sigma_\alpha$ and $R^2_{\text{ling}}$ alongside per-language
scores would make this distinction visible by default; current
Gini and max--min summaries cannot. The structure of the Bayesian decomposition
attributes is already visible non-parametrically via the statistical tests we perform.

\paragraph{Representation alignment to English is the most robust
single lever.}
The six standardized predictors entered together, only the
similarity between a model's hidden-state geometry for a language
and for English (\texttt{repr\_sim\_en}) is credibly positive in
every bucket: $+0.065$ on NLU, $+0.094$ on reasoning, and $+0.053$
on generation (Appendix~\ref{app:nlg}). A one-standard-deviation
gain is worth $6$--$9$ accuracy points. Two properties make this
useful. First, the predictor is a property of the
(model, language) pair, not of the language alone, which is why
its sign and magnitude carry across three task families scored
with three different metrics -- consistent with prior evidence
that decoder LLMs route multilingual computation through English
\citep{wendler-etal-2024-llamas, Li_Shi_Liu_Yang_Payani_Liu_Du_2025}.
Second, it is computable from a model alone, with no multilingual
benchmark required, making it the cheapest principled proxy
available for ranking checkpoints by expected multilingual
headroom. The coefficient is correlational, but its cross-bucket
stability is stronger evidence than any single-task association.

\paragraph{NLU and reasoning are not the same kind of problem.}
On NLU, model identity carries two-thirds of the total variance and
benchmark--model interaction is modest: a model strong on one
knowledge benchmark is strong on the next. On reasoning, the main
model effect drops to about a third of its NLU magnitude, and the
benchmark--model component carries nearly half of the total
variance, larger than the main model effect itself. Pooling
commonsense, math, and coreference into a single ``multilingual
reasoning'' score therefore averages over differences that exceed
the differences between models. Per-benchmark reporting is required,
and disagreement across reasoning benchmarks should be read as a
signal about the underlying tasks, not as measurement noise.

\paragraph{Fixed-effect-only specifications hide model-dependent
predictors.}
Tokenizer fertility is a good example. Its \emph{pooled} coefficient
on Reasoning is statistically null, averaged across all models;
fertility looks like it does not matter. But once we allow each
model its own fertility slope, that random slope accounts for
$12.1\%$ of total variance, with individual models' sensitivities
spread roughly $\pm 6$ accuracy points around the pooled mean. In other words, fertility hurts some models in some languages a lot,
and others barely at all, averaging cancels these opposing
effects out to zero. A fixed-effects-only analysis would have
wrongly concluded that fertility is inert; the random slope reveals
that \emph{which} languages a tokenizer over-fragments determines
\emph{which} languages a given model fails on, and the magnitude
of this effect is comparable to the main ``model'' effect on
Reasoning. This matches prior tokenizer findings
\citep{rust-etal-2021-good, ahia-etal-2023-languages}, and argues
for routinely reporting random slopes whenever an evaluation panel
mixes models with different tokenizer families.

\paragraph{The model--language residual sets the ceiling for
feature-based interventions.}
A $\mathrm{model} \times \mathrm{language}$ component of $7$--$9\%$
($\sigma \approx 0.044$--$0.053$) persists in both buckets after
every measured covariate. These are model-specific language
preferences of about $\pm 5$ accuracy points that none of the
linguistic features, the alignment probe, or the fertility slope
account for. They almost certainly trace to pre-training data
decisions -- mixture ratios, deduplication, quality filters --
that current model releases do not disclose. This residual
bounds what any language-feature-only intervention can close, and
it is the natural next target for the framework
\citep{blasi-etal-2022-systematic, blevins-zettlemoyer-2022-language}.

% \paragraph{The Bayesian decomposition is not a specification
% artifact.}
% Two distribution-free tests were reported before the Bayesian fit
% (Section~\ref{sec:nonparam}) make the same structural points that the
% mixed model later quantifies. The Friedman test rejects equal
% language ranks in $18/18$ benchmarks, with $W \in [0.31, 0.91]$
% and a median around $0.75$. The Kruskal--Wallis omnibus on resource
% tiers is rejected at $p < 10^{-15}$ in each
% $(\text{benchmark}, \text{task})$ cell, with all ten pairwise
% contrasts significant. The structure of the Bayesian decomposition
% attributes is already visible non-parametrically.

\paragraph{How this disparity can be addressed} Two of the variance components our framework isolates correspond to actionable levers. The first is representation alignment to English, raised either by cross-lingual alignment objectives at pretraining or by alignment fine-tuning afterward. The second is the residual model--language component -- the $7$--$9\%$ gap no observable language feature explains -- which our results trace to in-language pretraining data: mixture and quality. 

\section{Conclusion}
Our work reframes multilingual LLM evaluation from a passive mapping of performance gaps into an explainable, diagnostic framework: decomposing observed variance through a hierarchical model shows that most of the cross-lingual disparity is structural, attributable to predictable linguistic properties and internal representation geometries. Reporting $\sigma_\alpha$ alongside per-language accuracy attributes any reported gain to the component that actually moved; a model's internal representation alignment to English serves as a benchmark-free proxy for expected multilingual capacity during pretraining or alignment, and the residual model--language component sets the ceiling for feature-based interventions, pointing to pretraining-data mixture and curation as the next lever.

\section*{Limitations}
\label{sec:limitations}

Our analysis is observational. The seven-checkpoint panel is large
enough to identify the variance components reported in
Table~\ref{tab:variance} but small enough that the per-model random
slopes are aggressively shrunken by the prior; more checkpoints
would tighten both. The Gaussian likelihood on raw accuracy is an
approximation near the bounded scale. The
NLG bucket admits the same specification, but its twelve-language
Indic and Dravidian pool under-identifies the script and family
categoricals; we therefore restrict the NLG result to the single
\texttt{repr\_sim\_en} coefficient in Appendix~\ref{app:nlg} and
do not report a bucket-level $R^2_{\text{ling}}$ there. Finally,
all predictor coefficients are correlational: the cross-bucket
consistency of \texttt{repr\_sim\_en} is the strongest regularity
we identify, but causal claims require controlled training-side
experiments that we do not run here.

% % Bibliography entries for the entire Anthology, followed by custom entries
% % \bibliography{anthology,custom}
% % Custom bibliography entries only
% \bibliography{anthology-1, anthology-2, custom}
\bibliography{anthology-part-1, anthology-part-2, anthology-part-3, custom}

\appendix

\section{Appendix}

Table~\ref{tab:model_bucket_disparity} reports per-(model, bucket) disparity summaries --- mean score, Gini, Sen welfare, coefficient of variation, and relative and $z$-scaled max--min gaps --- aggregated across benchmarks within each bucket, providing the full numerical backing for the dispersion claims discussed in Section~\ref{sec:characterizing_disparity}. 

Table~\ref{tab:tier_mean_score_per_model} breaks the same scores down by resource tier and model, showing that the monotonic Class~1$\to$Class~5 staircase recurs for every checkpoint. 

Table~\ref{tab:kw_joshi_tiers} gives the Kruskal--Wallis omnibus statistics and tier-mean $z$-scores that motivate the Bayesian decomposition in Section~\ref{sec:whatexplains}; the corresponding full pairwise Dunn matrices appear in Appendix~\ref{app:dunn_full}. 

\begin{table*}[!htbp]
\centering
\small
\setlength{\tabcolsep}{5pt}
\renewcommand{\arraystretch}{1.15}
\begin{tabular}{lrrrrrl}
\toprule
\textbf{Benchmark} & $k$ (lang) & $m$ & $\chi^2$ & $p$ & Kendall $W$ & sig \\
\midrule
\rowcolor{bucketReason!55} \textsc{belebele}            & 62 & 7 & 297.1 & $3.4\!\times\!10^{-61}$ & 0.696 & *** \\
\rowcolor{bucketNLU!55}    \textsc{boolq-indic}         & 11 & 7 & 56.7  & $2.1\!\times\!10^{-10}$ & 0.810 & *** \\
\rowcolor{bucketNLU!55}    \textsc{global-mmlu}         & 15 & 7 & 71.4  & $2.1\!\times\!10^{-13}$ & 0.728 & *** \\
\rowcolor{bucketReason!55} \textsc{gsm8k-indic}         & 11 & 7 & 56.6  & $2.3\!\times\!10^{-10}$ & 0.808 & *** \\
\rowcolor{bucketNLU!55}    \textsc{include}             & 43 & 7 & 195.6 & $1.6\!\times\!10^{-39}$ & 0.665 & *** \\
\rowcolor{bucketReason!55} \textsc{mgsm}                & 11 & 7 & 39.3  & $6.1\!\times\!10^{-7}$  & 0.562 & *** \\
\rowcolor{bucketNLU!55}    \textsc{milu}                & 11 & 7 & 60.9  & $2.9\!\times\!10^{-11}$ & 0.871 & *** \\
\rowcolor{bucketNLU!55}    \textsc{mmlu-indic-roman}    & 10 & 7 & 57.1  & $1.7\!\times\!10^{-10}$ & 0.907 & *** \\
\rowcolor{bucketNLU!55}    \textsc{mmlu-prox}           & 29 & 7 & 124.1 & $2.2\!\times\!10^{-24}$ & 0.633 & *** \\
\rowcolor{bucketNLU!55}    \textsc{mmmlu}               & 14 & 7 & 73.7  & $7.3\!\times\!10^{-14}$ & 0.809 & *** \\
\rowcolor{bucketNLU!55}    \textsc{okapi-mmlu}          & 34 & 7 & 168.4 & $1.0\!\times\!10^{-33}$ & 0.729 & *** \\
\rowcolor{bucketNLU!55}    \textsc{triviaqa-indic-mcq}  & 11 & 7 & 60.2  & $4.0\!\times\!10^{-11}$ & 0.860 & *** \\
\rowcolor{bucketReason!55} \textsc{xcopa}               & 11 & 7 & 21.6  & $1.0\!\times\!10^{-3}$  & 0.309 & ** \\
\rowcolor{bucketReason!55} \textsc{xstorycloze}         & 11 & 7 & 40.5  & $3.7\!\times\!10^{-7}$  & 0.578 & *** \\
\rowcolor{bucketReason!55} \textsc{xwinograd}           &  6 & 7 & 27.2  & $1.3\!\times\!10^{-4}$  & 0.778 & *** \\
\bottomrule
\end{tabular}
\caption{Per-benchmark Friedman test (languages as items, $m{=}7$
models as judges) for the NLU and Reasoning buckets.
Kendall's $W=\chi^2/[m(k{-}1)]\in[0,1]$ is an effect-size analog.
Rows tinted by bucket
(\colorbox{bucketNLU}{NLU}, \colorbox{bucketReason}{Reasoning}).}
\label{tab:friedman_per_benchmark}
\end{table*}

\begin{table*}[!htbp]
\centering
\small
\setlength{\tabcolsep}{5pt}
\renewcommand{\arraystretch}{1.15}
\begin{tabular}{llcccccc}
\toprule
\textbf{Model} & \textbf{Bucket} &
$\mu$\,\hup & Gini\,\hdn & $W_{\text{Sen}}$\,\hup & CV\,\hdn &
$\Delta_{\text{rel}}$\,\hdn & $\Delta_{z}$\,\hdn \\
\midrule
\famrow{famQwen}   \textsc{Qwen3.5-122B-A10B} & NLU       & \textbf{0.741} & \textbf{0.064} & \textbf{0.695} & \textbf{0.099} & \textbf{0.363} & \textbf{3.581} \\
\famrow{famQwen}                              & Reasoning & 0.647          & 0.119          & 0.602          & 0.290          & 1.030          & 3.638 \\
\addlinespace[2pt]
\famrow{famQwen}   \textsc{Qwen3.5-4B}        & NLU       & 0.525          & 0.118          & 0.464          & 0.197          & 0.664          & 3.422 \\
\famrow{famQwen}                              & Reasoning & 0.567          & 0.134          & 0.506          & 0.273          & 0.890          & 3.075 \\
\addlinespace[2pt]
\famrow{famCohere} \textsc{aya-expanse-32b}   & NLU       & 0.562          & 0.112          & 0.502          & 0.180          & 0.599          & 3.338 \\
\famrow{famCohere}                            & Reasoning & \textbf{0.677} & 0.120          & 0.598          & 0.226          & 0.684          & \textbf{2.986} \\
\addlinespace[2pt]
\famrow{famGPT}    \textsc{gpt-oss-20b}       & NLU       & 0.440          & 0.134          & 0.391          & 0.185          & 0.721          & 3.753 \\
\famrow{famGPT}                               & Reasoning & 0.638          & 0.080          & 0.589          & 0.158          & 0.548          & 3.316 \\
\addlinespace[2pt]
\famrow{famSarvam} \textsc{sarvam-105b}       & NLU       & 0.591          & 0.108          & 0.539          & 0.171          & 0.630          & 3.705 \\
\famrow{famSarvam}                            & Reasoning & 0.648          & \textbf{0.075} & \textbf{0.599} & \textbf{0.143} & \textbf{0.481} & 3.165 \\
\addlinespace[2pt]
\famrow{famSarvam} \textsc{sarvam-30b}        & NLU       & 0.354          & 0.110          & 0.325          & 0.153          & 0.640          & 3.939 \\
\famrow{famSarvam}                            & Reasoning & 0.379          & 0.131          & 0.345          & 0.247          & 0.754          & 3.213 \\
\addlinespace[2pt]
\famrow{famCohere} \textsc{tiny-aya-global}   & NLU       & 0.421          & 0.134          & 0.368          & 0.213          & 0.712          & 3.363 \\
\famrow{famCohere}                            & Reasoning & 0.525          & 0.143          & 0.464          & 0.278          & 0.882          & 3.026 \\
\bottomrule
\end{tabular}
\caption{Per-(model, bucket) language disparity, unweighted means
across benchmarks in the bucket
(\colorbox{bucketNLU}{NLU}, \colorbox{bucketReason}{Reasoning}).
For each $(\text{model},\text{benchmark})$ cell, metrics are
computed over the per-language scores $\{x_1,\dots,x_L\}$ and then
averaged across benchmarks within the bucket.
\textbf{$\mu$}~$=\frac{1}{L}\sum_i x_i$ is the mean per-language
score (higher is better).
\textbf{Gini}~$=\frac{\sum_i\sum_j |x_i-x_j|}{2L^2\mu}\in[0,1]$
measures relative inequality across languages (lower is more
equal).
\textbf{$W_{\text{Sen}}$}~$=\mu\,(1-\text{Gini})$ is Sen's
welfare index, an equity-discounted mean (higher is better).
\textbf{CV}~$=\sigma/\mu$ is the coefficient of variation, a
scale-free spread (lower is better).
\textbf{$\Delta_{\text{rel}}$}~$=(\max_i x_i-\min_i x_i)/\mu$ is
the best-vs.-worst gap normalised by the mean (lower is better).
\textbf{$\Delta_{z}$}~$=(\max_i x_i-\min_i x_i)/\sigma$ is the
same gap expressed in standard deviations (lower is better).
Arrows show direction of improvement; bold marks the best value per
column within each bucket.}
\label{tab:model_bucket_disparity}
\end{table*}

\begin{table*}[!htbp]
\centering
\small
\setlength{\tabcolsep}{6pt}
\renewcommand{\arraystretch}{1.15}
\begin{tabular}{lccccc}
\toprule
\textbf{Model} & \textbf{1-Scraping}\,\hup & \textbf{2-Hopefuls}\,\hup & \textbf{3-Rising}\,\hup & \textbf{4-Underdogs}\,\hup & \textbf{5-Winners}\,\hup \\
\midrule
\famrow{famQwen}   \textsc{Qwen3.5-122B-A10B} & \textbf{0.513} & \textbf{0.620} & \textbf{0.684} & \textbf{0.793} & 0.806          \\
\famrow{famQwen}   \textsc{Qwen3.5-4B}        & 0.361          & 0.413          & 0.498          & 0.669          & 0.735          \\
\famrow{famCohere} \textsc{aya-expanse-32b}   & 0.473          & 0.482          & 0.646          & 0.694          & \textbf{0.808} \\
\famrow{famGPT}    \textsc{gpt-oss-20b}       & 0.459          & 0.500          & 0.510          & 0.558          & 0.680          \\
\famrow{famSarvam} \textsc{sarvam-105b}       & 0.514          & 0.600          & 0.610          & 0.617          & 0.747          \\
\famrow{famSarvam} \textsc{sarvam-30b}        & 0.244          & 0.355          & 0.320          & 0.350          & 0.448          \\
\famrow{famCohere} \textsc{tiny-aya-global}   & 0.288          & 0.371          & 0.453          & 0.551          & 0.658          \\
\bottomrule
\end{tabular}
\caption{Mean per-language score by resource tier for each
model, averaged across the NLU and Reasoning buckets. Construction
matches Table~\ref{tab:tier_mean_score}. Bold marks the best model
per tier.}
\label{tab:tier_mean_score_per_model}
\end{table*}

\begin{table*}[!htbp]
\centering
\small
\setlength{\tabcolsep}{5pt}
\renewcommand{\arraystretch}{1.15}
\begin{tabular}{lcccccc}
\toprule
\textbf{Bucket} & $H$ & $p$ & C1 & C2 & C3 & C4\,/\,C5 \\
\midrule
\cellcolor{bucketNLU!55}NLU       & 1243.89 & $4.9\!\times\!10^{-268}$ & $-1.02$ & $-0.40$ & $-0.01$ & $+0.21\,/\,+0.75$ \\
\cellcolor{bucketReason!55}Reasoning & 78.69   & $3.3\!\times\!10^{-16}$  & $-0.78$ & $-0.32$ & $-0.04$ & $+0.24\,/\,+0.81$ \\
\bottomrule
\end{tabular}
\caption{Kruskal--Wallis omnibus across the five resource tiers on
$z$-normalised scores (within each $(\text{benchmark},
\text{task\_id})$ cell), with tier-mean $z$-scores from Class~1 to
Class~5. Dunn pairwise tests with BH--FDR correction: all $10$
contrasts significant in both buckets (weakest C1 vs.\ C2 in
Reasoning at $p{=}0.014$). The same staircase recurs per model.}
\label{tab:kw_joshi_tiers}
\end{table*}

\section{Design choices: covariates, random structure, and probe corpus}
\label{app:design_choices}

The methodology paragraph in
Section~\ref{sec:disparity:method} compresses several independent
design choices into a single line. We expand each here, stating
for each what we use, why, what alternatives we considered, and
where the choice is standard in the literature.

\subsection{Why hierarchical Bayesian modeling}

Our data has the structure of a crossed panel: every
$(\text{model}, \text{benchmark}, \text{language})$ cell yields one
accuracy observation, and the three index sets are non-nested --
the same language appears under every model and every benchmark,
the same model appears under every benchmark and every language.
Three features of this setting drive the modeling choice.

First, the quantity of interest is not a single point estimate but
a \emph{variance decomposition}: how much of the cross-language
dispersion is attributable to the language itself, to its
interaction with a specific model, and with its interaction with a
specific benchmark. Random-effect intercepts give each of these
components a single scalar ($\sigma_\alpha$,
$\sigma_{\mathrm{model:lang}}$, $\sigma_{\mathrm{bench:lang}}$) on
the accuracy scale, which is exactly the quantity the headline
$R^2_{\text{ling}}$ statistic compares across model stages. An OLS
regression with the same covariates plus a per-cell fixed effect,
or an ANOVA-style sums-of-squares decomposition, would report
variance \emph{explained} rather than variance \emph{components},
and neither shrinks sparsely-observed cells.

Second, several of the crossings are sparsely observed: many
language--benchmark cells have a single trial. Maximum-
likelihood mixed-effect estimators (e.g.\ \texttt{lme4},
\texttt{glmmTMB}) frequently fail to converge under the full
random-effect structure at this sparsity, and a fully fixed-effect
specification overfits the smallest cells. Hierarchical Bayesian
estimation with weakly informative priors regularises these toward
their group mean and returns a coherent joint posterior over the
variance components, which is what the credibility intervals on
$R^2_{\text{ling}}$ are computed from; a frequentist bootstrap on
the same components does not yield this joint distribution.

Third, the framework is the standard analytic tool for crossed
language/item/subject panels in psycholinguistics
\citep{baayen-etal-2008-mixed}, has been explicitly recommended for
LLM-evaluation panels by \citet{ulmer-etal-2022-experimental}, and is
used in this form for multilingual benchmark analysis by
\citep{khanuja-etal-2023-evaluating, ahuja-etal-2023-mega,
ojo-etal-2025-afrobench}. We follow their specification.

\begin{table}[!htbp]
\centering
\small
\setlength{\tabcolsep}{4pt}
\renewcommand{\arraystretch}{1.15}
\begin{tabularx}{\columnwidth}{@{}
  >{\raggedright\arraybackslash}p{0.23\columnwidth}
  >{\raggedright\arraybackslash}X
  >{\raggedright\arraybackslash}p{0.21\columnwidth}@{}}
\toprule
\textbf{Feature} & \textbf{Source} & \textbf{Type} \\
\midrule
\multicolumn{3}{l}{\textit{Language-level features}} \\[-2pt]
\rowcolor{featLang!55}
\texttt{script}         & Hardcoded ISO-1 $\rightarrow$ script map                                                       & Categorical (15 levels)         \\
\rowcolor{featLang!55}
\texttt{family}         & Hardcoded ISO-1 $\rightarrow$ family map                                                       & Categorical (17 levels)         \\
\rowcolor{featLang!55}
\texttt{resource\_class}   & \citet{joshi-etal-2020-state} resource class                                                   & Integer 1--5 $\rightarrow$ std. \\
\rowcolor{featLang!55}
\texttt{syn\_dist\_en}  & \texttt{lang2vec} syntactic distance to English                                                & Float $\rightarrow$ std.        \\
\rowcolor{featLang!55}
\texttt{phon\_dist\_en} & \texttt{lang2vec} phonological distance to English                                             & Float $\rightarrow$ std.        \\[4pt]
\multicolumn{3}{l}{\textit{Model $\times$ language features}} \\[-2pt]
\rowcolor{featModel!55}
\texttt{fertility}      & \texttt{compute\_fertility} — tokens-per-word for the model's tokenizer on a language sample & Float $\rightarrow$ std.        \\
\rowcolor{featModel!55}
\texttt{repr\_sim\_en}  & \texttt{compute\_repr\_sim} — cosine similarity of hidden-state geometry for a language to English & Float $\rightarrow$ std.   \\
\bottomrule
\end{tabularx}
\caption{Features used per language and model, grouped by scope
(\colorbox{featLang}{language-level},
\colorbox{featModel}{model $\times$ language}).
All continuous features are standardized ($\mu=0$, $\sigma=1$) before use.}
\label{tab:features}
\end{table}

\section{Full model specification}
\label{app:model}

\begin{figure*}[!htbp]
\centering
\begin{minipage}{0.95\textwidth}
\textbf{Stage 1 --- Baseline} (used to estimate the raw cross-language gap $\sigma_\alpha$).
\begin{Verbatim}[fontsize=\small,frame=single,framesep=4pt,breaklines=true,breakanywhere=true]
score ~ 1 + C(task) + (1|language) + (1|model)
\end{Verbatim}

\vspace{0.4em}
\textbf{Stage 2 --- Language-features fit} (used to compute $R^2_{\text{ling}}$ via Equation~\ref{eq:r2ling}).
\begin{Verbatim}[fontsize=\small,frame=single,framesep=4pt,breaklines=true,breakanywhere=true]
score ~ 1 + joshi_class_z + syn_dist_en_z + phon_dist_en_z + geo_dist_en_z
          + C(script) + C(family) + C(task)
          + (1|language) + (1|model)
\end{Verbatim}

\vspace{0.4em}
\textbf{Stage 3 --- Full model} (used for all coefficient and variance-share tables in the main text).
\begin{Verbatim}[fontsize=\small,frame=single,framesep=4pt,breaklines=true,breakanywhere=true]
score ~ 1 + joshi_class_z + syn_dist_en_z + phon_dist_en_z + geo_dist_en_z
          + C(script) + C(family) + C(task)
          + fertility_z + repr_sim_en_z
          + (1|language) + (1|model)
          + (1|model:language) + (1|task:language) + (1|task:model)
          + (0 + fertility_z|model)
\end{Verbatim}
\end{minipage}
\caption{The three nested model specifications fit in this work, in
\texttt{bambi}/\texttt{lme4} formula syntax. Stage~2 adds structured
language-level covariates on top of the Stage~1 baseline; Stage~3
adds model-conditional covariates (\texttt{fertility\_z},
\texttt{repr\_sim\_en\_z}), two-way random interactions, and a
per-model random slope on fertility.}
\label{fig:model_specs}
\end{figure*}

All fits share a common formula template, with the right-hand side
extended in two steps from the baseline. We write the model in
\texttt{bambi}/\texttt{lme4} formula syntax; the three stages are
shown together in Figure~\ref{fig:model_specs}.

\paragraph{Priors.}
We use \texttt{bambi}'s defaults throughout: weakly-informative Normal
priors on fixed effects (centred at $0$ with prior SD scaled to the
predictor), Half-Normal priors on random-effect SDs, and Half-StudentT
on the residual SD. Standardised continuous predictors carry a
$\mathcal{N}(0,1)$ prior on the standardised scale.

\paragraph{Sampling.}
NUTS via \texttt{numpyro} (JAX-backed), four chains with
\texttt{chain\_method="vectorized"} on a single A100, $5000$ warmup and
$2000$ post-warmup draws per chain, target acceptance $0.99$.
Convergence is assessed by $\hat{R}$ and bulk effective sample size.

\subsection{Language-level covariates}

\paragraph{Resource class.}
We use the six-tier taxonomy ($0$--$5$) of
\citet{joshi-etal-2020-state}, treated as a standardised continuous
covariate. It is a single scalar that captures the NLP-resource
situation of a language (availability of labelled data, raw text,
and tools) more directly than raw demographics. Among the
alternatives considered -- log token count from CommonCrawl, mC4,
or OSCAR; Wikipedia article count; and Ethnologue speaker
population -- each conflates the quantity of interest with web-crawl
artefacts, encyclopaedic activity, or user demand respectively. The
resource taxonomy is the de facto resource proxy in multilingual NLP
and is used directly in
\citep{khanuja-etal-2023-evaluating, ojo-etal-2025-afrobench, ahuja-etal-2023-mega} among others.

\paragraph{Typological distances to English.}
We use the syntactic, phonological, and geographic distances from
\texttt{lang2vec}/URIEL \citep{littell-etal-2017-uriel}, each
standardised before entering the regression. These are the three
least mutually redundant distance families in URIEL: syntactic
distance summarises word-order and morphology, phonological
distance summarises sound inventory (relevant to subword
segmentation), and geographic distance captures areal/contact
effects orthogonal to genealogy. We exclude the URIEL
\emph{genetic} distance because it duplicates the categorical
\texttt{family} factor we include separately. We considered raw
WALS features, Glottolog family identifiers only, 
but rejected them due to lack of a continuous distance, and incomplete
language coverage respectively. URIEL/lang2vec is the dominant
cross-lingual distance toolkit in multilingual transfer
\citep{lauscher-etal-2020-zero}.

\paragraph{Script and family as categorical factors.}
We include $\mathrm{script}(\ell)$ and $\mathrm{family}(\ell)$ as
unordered categorical factors with a coarse grouping ($7$ scripts;
$8$ families on our panel). Categorical coding is the natural
choice as scripts (Latin, Cyrillic, Devanagari, Arabic, CJK,
Ge\textquotesingle ez, Brahmic-other) and families (Indo-European,
Sino-Tibetan, Afro-Asiatic, Niger-Congo, Dravidian, Turkic,
Austronesian, Uralic) admit no defensible ordering. Despite the
URIEL distances already encoding typology, we keep both:
$\mathrm{script}$ captures engineering-level effects (subword
vocabulary coverage, romanisation availability) that no continuous
distance reflects, and $\mathrm{family}$ captures pretraining-data
clustering, since languages of the same family typically co-occur
in web crawls. We considered a binary Latin / non-Latin coding and
a per-script random intercept; the former collapses Devanagari,
Arabic, and CJK into one bucket and hides script-specific effects
we observe empirically, while the latter is not data-efficient at
our $7$-script panel.

\subsection{Model--language covariates}

\paragraph{Tokenizer fertility.}
We define fertility as the mean number of subword tokens per
whitespace-separated word, computed on FLORES-200 (see corpus
discussion below). Fertility is the most widely used
tokenizer-quality proxy in multilingual NLP and has documented
downstream effects on both accuracy and cost
\citep{rust-etal-2021-good, ahia-etal-2023-languages, petrov-etal-2023-language}.
We considered bytes-per-token and the tokenizer parity ratio
\citep{petrov-etal-2023-language}, and the inverse characters-per-token
compression rate; we chose fertility because it preserves the
subword-fragmentation signal that affects context utilisation,
avoids the reference-language dependency of a parity ratio, and
follows the definition used by
\citep{rust-etal-2021-good, ahia-etal-2023-languages}.

\paragraph{Representation similarity to English (\texttt{repr\_sim\_en}).}
We compute Centered Kernel Alignment
\citep{kornblith19a} between the hidden-state matrices of an
English FLORES-200 sentence and its translation in the target
language, at the model's middle layer
($\lfloor L/2 \rfloor$ for an $L$-layer model), mean-pooled across
content tokens. Cross-lingual hidden-state comparison requires a
metric invariant to orthogonal rotation and isotropic scaling of
the representation space and well-defined between matrices of
possibly different dimension; CKA is the dominant tool that meets
all three requirements and is standard in LLM-internals analyses
\citep{kornblith19a, wendler-etal-2024-llamas, dumas-etal-2025-separating, Li_Shi_Liu_Yang_Payani_Liu_Du_2025}.
We considered mean cosine similarity over paired tokens, SVCCA
\citep{raghu-etal-2017-svcca}, Representational Similarity Analysis,
and Procrustes alignment; we rejected them respectively because
they are not rotation-invariant, require matched dimensions and
are unstable at our sample sizes, offer only rank-level invariance,
and depend on a paired token-level alignment that is ill-defined
when source and target tokenisations differ.
We summarise at the middle layer because decoder-only LLMs exhibit
a U-shaped language-specificity profile -- early layers dominated
by surface token form, late layers by output-language reformatting,
and mid-depth the most language-agnostic
\citep{wendler-etal-2024-llamas, dumas-etal-2025-separating, Li_Shi_Liu_Yang_Payani_Liu_Du_2025}. We mean-pool because the
source and target tokenisations differ token-by-token, ruling out
positional alignment; mean-pooling is the standard choice in
CKA-on-LLM analyses, whereas last-token pooling is unstable on
short FLORES-200 sentences.

\subsection{Random-effect structure}

\paragraph{Three two-way interactions, no three-way.}
We include
$(1 \mid \mathrm{model}{:}\mathrm{language})$,
$(1 \mid \mathrm{bench}{:}\mathrm{language})$, and
$(1 \mid \mathrm{bench}{:}\mathrm{model})$ as random intercepts.
The three-way
$(\mathrm{model}{:}\mathrm{bench}{:}\mathrm{language})$
interaction is omitted because with one observation per
$(\text{model}, \text{benchmark}, \text{language})$ cell it is
unidentified from the residual $\varepsilon$ and would simply
re-label noise. We use random rather than fixed interactions
because a panel of $7$ models, $\sim 18$ benchmarks, and $54$
languages yields hundreds of pairwise interactions: a fixed-effect
specification would overfit and produce unstable per-cell
estimates, whereas random-effect shrinkage regularises toward
zero where data is sparse. Crossed-random-effects models of this
form are standard in psycholinguistics
\citep{baayen-etal-2008-mixed} and are explicitly recommended for
LLM-evaluation panels by \citet{ulmer-etal-2022-experimental}.

\paragraph{Per-model random slope on fertility.}
We add $(0 + \texttt{fertility\_z} \mid \mathrm{model})$ -- one
fertility coefficient per model -- but no analogous slopes for the
other covariates. Prior work
\citep{rust-etal-2021-good, ahia-etal-2023-languages} documents
that tokenizers from different families penalise different
languages, so the per-model fertility effect is expected to vary;
a pooled fixed effect averages these opposing slopes toward zero,
which is what we observe (Section~\ref{sec:disparity:variance}).
The other covariates are properties of the language alone
($\mathbf{x}_\ell$), so per-model slopes would imply ``each model
has its own typology'', which has no mechanistic justification and
did not improve leave-one-language-out predictive accuracy in
pilot fits (within $\pm 1$ ELPD-SE). A
$\texttt{fertility\_z} \times \texttt{model\_id}$ fixed interaction
is equivalent to a random slope with a flat prior and is less
stable at our sample size.

\subsection{Measurement corpus: FLORES-200}

We compute both fertility and \texttt{repr\_sim\_en} on the
FLORES-200 \texttt{devtest} split \citep{Costa-jussa2024}. A parallel
corpus is required because cross-lingual hidden-state comparison
must hold semantic content fixed -- otherwise CKA conflates
language-geometry differences with content differences -- and
fertility is comparable across languages only when measured on the
same underlying content. We chose FLORES-200 because it is
professionally human-translated across $200$ languages, fully
covers all $54$ languages in our model panel without imputation,
is domain-consistent (Wikipedia news/encyclopedia), and is the
standard parallel evaluation corpus in multilingual LLM work
\citep{Costa-jussa2024}. We considered Tatoeba,
OPUS / mC4 parallel subsets, and the NLLB seed corpus, but each
is weaker on coverage, translation quality, or per-language size,
the last of which would inflate CKA estimator variance.

\section{Per-feature coefficients of the full model}
\label{app:coefs}

The main text (Section~\ref{sec:disparity:r2ling}) reports the
headline finding that \texttt{repr\_sim\_en} is the only continuous
predictor credibly non-null in both buckets. This appendix presents
the full per-feature picture numerically (Table~\ref{tab:coefs}),
and discusses why the four non-named predictors are individually
attenuated despite the high joint $R^2_{\text{ling}}$.

\paragraph{Resource level (\texttt{resource\_class}) is credible on
Reasoning and borderline on NLU.}
Resource class is credibly positive on Reasoning ($+0.037$ $[0.002, 0.07]$)
and just-crosses-zero on NLU ($+0.022$ $[-0.005, 0.05]$). The full-model
coefficient understates the marginal effect: in the
language-features-only fit (without \texttt{repr\_sim\_en}),
$\beta_{\mathrm{resource}}$ rises to $+0.038$ on NLU and $+0.057$ on
Reasoning, both then credibly positive. The reduction in the full
model is expected --- resource level is itself a coarse predictor of
representation alignment, and the two compete for shared variance.

\paragraph{Why the four typology / fertility predictors are not
individually credible.}
Syntactic, phonological, and geographic distance to English, together
with tokenizer fertility, all have HDIs that span zero in both
substantive buckets. This is the typical pattern when correlated
predictors are entered jointly into a mixed model:
\texttt{syn\_dist\_en} and \texttt{phon\_dist\_en} are themselves
correlated with each other and with the categorical
$\mathrm{family}(\ell)$ term, while \texttt{geo\_dist\_en} (an areal
proxy) overlaps with both \texttt{syn\_dist\_en} and the script
categoricals. The language random intercept further absorbs
residual language-level structure. The pattern is collinear shrinkage
of individual coefficients, not absence of feature signal: the joint
$R^2_{\text{ling}}$ remains high because the features are
\emph{collectively} explanatory.

Fertility behaves differently again. Its pooled fixed effect is
washed out because its sign and magnitude vary by model, but the
random per-model slope $(0+\texttt{fertility\_z}\mid\mathrm{model})$
carries $12.1\%$ of Reasoning variance. Fertility \emph{does} matter, but
model-specifically rather than universally. We caution against
reading the borderline pooled HDIs as ``the feature is
uninformative'': for the typology predictors the right reading is
collective rather than individual explanatory power; for fertility
it is the random slope, not the fixed slope, that carries the
signal.

\begin{table}[!htbp]
\centering
\small
\setlength{\tabcolsep}{5pt}
\renewcommand{\arraystretch}{1.12}

\resizebox{\columnwidth}{!}{%
\begin{tabular}{lcc}
\toprule
\textbf{Predictor} &
\cellcolor{bucketNLU!55}\textbf{NLU} &
\cellcolor{bucketReason!55}\textbf{Reasoning} \\
\midrule

\texttt{resource\_class\_z}
  & $+0.022^{\dagger}$ & $+0.037^{\star}$ \\
  & {\scriptsize $[-0.005, +0.050]$}
  & {\scriptsize $[+0.002, +0.070]$} \\

\texttt{syn\_dist\_en\_z}
  & $-0.012$ & $+0.025$ \\
  & {\scriptsize $[-0.045, +0.020]$}
  & {\scriptsize $[-0.015, +0.070]$} \\

\texttt{phon\_dist\_en\_z}
  & $+0.009$ & $+0.007$ \\
  & {\scriptsize $[-0.004, +0.020]$}
  & {\scriptsize $[-0.012, +0.030]$} \\

\texttt{geo\_dist\_en\_z}
  & $-0.011$ & $-0.007$ \\
  & {\scriptsize $[-0.041, +0.020]$}
  & {\scriptsize $[-0.044, +0.030]$} \\

\texttt{fertility\_z}
  & $-0.021^{\dagger}$ & $-0.051^{\dagger}$ \\
  & {\scriptsize $[-0.050, +0.005]$}
  & {\scriptsize $[-0.112, +0.007]$} \\

\texttt{repr\_sim\_en\_z}
  & $+0.065^{\star}$ & $+0.094^{\star}$ \\
  & {\scriptsize $[+0.051, +0.079]$}
  & {\scriptsize $[+0.074, +0.116]$} \\

\bottomrule
\end{tabular}%
}

\caption{Posterior mean and $90\%$ highest density interval (HDI) of each
standardised continuous predictor in the full model for the NLU and Reasoning
buckets. For each predictor, the first row reports the posterior mean and the
second row reports the corresponding $90\%$ HDI in brackets. Markers:
$\star$ = HDI excludes zero; $\dagger$ = HDI just crosses zero at one boundary.}
\label{tab:coefs}
\end{table}

% Variable glossary for the Bayesian hierarchical framework.
% Full-width longtable; spans both columns by switching to single-column
% layout for the duration of the table. Place inside the Bayesian-modelling
% appendix with \input{tables/variable_glossary.tex}.
\onecolumn
{\small
\renewcommand{\arraystretch}{1.25}
\setlength{\tabcolsep}{4pt}
\begin{longtable}{@{}p{0.16\textwidth} p{0.30\textwidth} p{0.50\textwidth}@{}}
\toprule
\textbf{Symbol / name} & \textbf{Formula term / source} & \textbf{Definition} \\
\midrule
\endfirsthead

\multicolumn{3}{c}{\textit{Table (continued).}}\\
\toprule
\textbf{Symbol / name} & \textbf{Formula term / source} & \textbf{Definition} \\
\midrule
\endhead

\bottomrule
\multicolumn{3}{r}{\textit{Continued on next page.}}\\
\endfoot

\bottomrule
\caption{Glossary of every variable in the Bayesian hierarchical
framework. Symbols and \texttt{bambi}/\texttt{lme4} formula names are
aligned with Sections~\ref{sec:disparity:method},
\ref{sec:disparity:r2ling} and the modelling appendix.
``$\ell$'' indexes language, ``$m$'' model, ``$t$'' benchmark/task,
and ``$c = (m, t, \ell)$'' a single observation cell.}
\label{tab:var_glossary}\\
\endlastfoot

% ---------------- Indices and observation unit ----------------
\multicolumn{3}{@{}l}{\textbf{Indices and Observation Unit}}\\[2pt]
$m$ & Model identifier (7 models). & One of the seven evaluated LLMs (e.g., \textsc{Qwen-3.5-122B}, \textsc{aya-expanse-32b}, \ldots; see Table~\ref{tab:models} ). \\
$\ell$ & Language identifier. & ISO-1 code. Per-bucket pools contain $63$ languages for NLU, $62$ for Reasoning, and $12$ for NLG, restricted to languages with both \texttt{fertility} and \texttt{repr\_sim\_en} available. \\
$t$ & Benchmark or task identifier. & After benchmark-as-task aggregation, this equals the benchmark name. Translation direction is preserved as a separate task in NLG. Counts: $9$ for NLU, $6$ for Reasoning, $5$ for NLG. \\
$c = (m, t, \ell)$ & Observation cell. & A single aggregated row in the fit frame, expanded to $(m, t, \ell, \text{direction})$ for NLG translation. \\
$y_c$, \texttt{score} & Outcome in $[0,1]$. & Multiple-choice accuracy (NLU and most Reasoning tasks), \texttt{exact\_match,flexible-extract} (GSM variants), or \texttt{chrF}/$100$ (NLG). \\
\midrule

% ---------------- Language-level features ----------------
\multicolumn{3}{@{}l}{\textbf{Language-Level Features} (functions of $\ell$ only)}\\[2pt]
\texttt{resource\_class}, \texttt{resource\_class\_z} & Integer $1$--$5$; z-scored. & Resource class from \citet{joshi-etal-2020-state}, standardised within each bucket fit frame. \\
\texttt{syn\_dist\_en}, \texttt{syn\_dist\_en\_z} & \texttt{lang2vec} Hamming distance on \texttt{syntax\_knn}. & Syntactic distance to English from URIEL \citep{littell-etal-2017-uriel}, standardised. \\
\texttt{phon\_dist\_en}, \texttt{phon\_dist\_en\_z} & \texttt{lang2vec} Hamming distance on \texttt{phonology\_knn}. & Phonological distance to English. \\
\texttt{geo\_dist\_en}, \texttt{geo\_dist\_en\_z} & Great-circle distance between \texttt{lang2vec} geographic centroids. & Areal/geographic distance to English used in this run. \\
$\mathrm{script}(\ell)$ & \texttt{C(script)} categorical. & Writing system. The reference level is absorbed into $\beta_0$. NLU has approximately $18$ levels; NLG has $10$, of which $9$ are singletons. \\
$\mathrm{family}(\ell)$ & \texttt{C(family)} categorical. & Language family. NLU has approximately $19$ levels; NLG has $2$ (Indo-Aryan and Dravidian). \\
$\mathbf{x}_\ell$ & Vector. & Stack of the four standardised numeric language features together with the script and family contrasts. \\
\midrule

% ---------------- Model x language features ----------------
\multicolumn{3}{@{}l}{\textbf{Model$\,\times\,$Language Features} (functions of $(m, \ell)$)}\\[2pt]
\texttt{fertility}, \texttt{fertility\_z} & Tokens per word for $m$'s tokeniser on language $\ell$. & Tokeniser inefficiency for $\ell$, standardised within bucket. Paired models share tokenisers (Qwen-122B/4B; sarvam-105b/30b). \\
\texttt{repr\_sim\_en}, \texttt{repr\_sim\_en\_z} & CKA similarity \citep{kornblith19a} between $m$'s middle-layer mean-pooled hidden states for $\ell$ and English. & Representation alignment to English; the single most consistent predictor across all three buckets. \\
$\mathbf{z}_{m,\ell}$ & Vector. & Stack of \texttt{fertility\_z} and \texttt{repr\_sim\_en\_z} evaluated at $(m, \ell)$. \\
\midrule

% ---------------- Fixed effects ----------------
\multicolumn{3}{@{}l}{\textbf{Fixed Effects} (population-level coefficients)}\\[2pt]
$\beta_0$ & \texttt{Intercept}, \texttt{1}. & Grand mean at the reference levels of all categorical variables. \\
$\tau_t$ & \texttt{C(task)}, one per benchmark. & Benchmark fixed effect, carried over from Eq.~\ref{eq:base}. \\
$\gamma_{\mathrm{resource}}$ & \texttt{resource\_class\_z}. & Effect of one standard deviation on the resource axis. \\
$\gamma_{\mathrm{syn}}$, $\gamma_{\mathrm{phon}}$, $\gamma_{\mathrm{geo}}$ & \texttt{syn\_/phon\_/geo\_dist\_en\_z}. & Effects of typological and areal distance to English. \\
$\gamma_{\mathrm{fert}}$ & \texttt{fertility\_z}. & Pooled fertility effect; per-model deviations are carried by $r_m$ below. \\
$\gamma_{\mathrm{rs}}$ & \texttt{repr\_sim\_en\_z}. & Representation-alignment effect; the $90\%$ HDI excludes zero in every bucket. \\
$\boldsymbol{\beta}^{\mathrm{script}}_{\!s}$, $\boldsymbol{\beta}^{\mathrm{family}}_{\!f}$ & \texttt{C(script)}, \texttt{C(family)}. & Per-level contrasts against the reference script or family. \\
$\boldsymbol{\beta}$, $\mathbf{x}_i$ & Stacked. & The term $\mathbf{x}_i^{\top}\boldsymbol{\beta}$ in Eq.~\ref{eq:full} bundles all of the above continuous and categorical fixed effects for cell $i$. \\
\midrule

% ---------------- Random effects ----------------
\multicolumn{3}{@{}l}{\textbf{Random Effects} (group-level intercepts and slopes)}\\[2pt]
$\alpha_\ell$ & \texttt{(1|language)}. & Language random intercept capturing residual language signal beyond features and categoricals, with $\alpha_\ell \sim \mathcal{N}(0, \sigma_\alpha^2)$. \\
$u_m$ & \texttt{(1|model)}. & Model random intercept, with $u_m \sim \mathcal{N}(0, \sigma_u^2)$. \\
$v_{m,\ell}$ & \texttt{(1|model:language)}. & Per-(model, language) idiosyncratic boost or deficit beyond the main effects. \\
$w_{t,\ell}$ & \texttt{(1|bench:language)}. & Per-(benchmark, language) difficulty quirk. \\
$s_{t,m}$ & \texttt{(1|bench:model)}. & Per-(benchmark, model) capability quirk; the dominant variance share on Reasoning ($46\%$). \\
$r_m$ & \texttt{(0 + fertility\_z|model)}. & Per-model random slope on fertility, allowing fertility's effect to vary across models. \\
$\varepsilon_c$ & \texttt{sigma}. & Gaussian residual, with $\varepsilon_c \sim \mathcal{N}(0, \sigma^2)$. \\
\midrule

% ---------------- Variance components ----------------
\multicolumn{3}{@{}l}{\textbf{Variance Components} (standard deviations of the random effects)}\\[2pt]
$\sigma_\alpha$ & Standard deviation of $\alpha_\ell$ in Stage~3 (or any single fit). & Size of the cross-language gap. Reported as $0.078$ on NLU and $0.088$ on Reasoning. \\
$\sigma_\alpha^{(\mathrm{lang})}$ & Standard deviation of $\alpha_\ell$ in Stage~2. & Residual language SD after the language covariates are added; the numerator of $R^2_{\text{ling}}$. \\
$\sigma_u$ & Standard deviation of $u_m$. & Drives the \texttt{model} variance share. \\
$\sigma_v$ & Standard deviation of $v_{m,\ell}$. & The \texttt{model$\times$language} variance share. \\
$\sigma_w$ & Standard deviation of $w_{t,\ell}$. & The \texttt{task$\times$language} variance share. \\
$\sigma_{tm}$ & Standard deviation of $s_{t,m}$. & The \texttt{task$\times$model} variance share. \\
$\sigma_{fm}$ & Standard deviation of $r_m$. & The \texttt{fertility-slope$\times$model} variance share ($12.1\%$ on Reasoning). \\
$\sigma$ & Residual standard deviation. & Within-cell noise. \\
\midrule

% ---------------- Derived / reported quantities ----------------
\multicolumn{3}{@{}l}{\textbf{Derived Quantities Reported in the Paper}}\\[2pt]
$R^2_{\text{ling}}$ & $1 - \bigl[\sigma_\alpha^{(\mathrm{lang})}\bigr]^2 / \sigma_\alpha^2$ \quad (Eq.~\ref{eq:r2ling}). & Proportional reduction in the language random-intercept variance attributable to observable language covariates. Reported as the posterior mean with a $90\%$ HDI. \\
\midrule

% ---------------- Equation-level objects ----------------
\multicolumn{3}{@{}l}{\textbf{Composite Objects Appearing in Equations~\ref{eq:base}--\ref{eq:full}}}\\[2pt]
$\mu$ & Intercept in Eq.~\ref{eq:base}. & Plays the same role as $\beta_0$ in Eq.~\ref{eq:full}: the global mean. \\
$\eta_c$ & Linear predictor. & $\mu + \tau_t + \mathbf{x}_i^{\!\top}\boldsymbol{\beta} + \alpha_\ell + u_m + v_{m,\ell} + w_{t,\ell} + s_{t,m}$ (Stage~3 full model). \\
$\mathrm{Stage~1}$ & Baseline. & Drops the structured-features block, keeping $\mu + \tau_t + \alpha_\ell + u_m + \varepsilon$. Estimates the raw $\sigma_\alpha$. \\
$\mathrm{Stage~2}$ & Language-features fit. & Adds $\mathbf{x}_\ell$ (the four $\gamma$ terms together with $\boldsymbol{\beta}^{\mathrm{script}}$ and $\boldsymbol{\beta}^{\mathrm{family}}$). Yields $\sigma_\alpha^{(\mathrm{lang})}$. \\
$\mathrm{Stage~3}$ & Full model. & Adds $\mathbf{z}_{m,\ell}$, the random interactions $(v, w, s)$, and the per-model fertility slope $r_m$. Used for all coefficient and variance-share tables. \\

\end{longtable}
}
\twocolumn

\section{Non-parametric cross-check of the feature ranking}
\label{sec:tree_validation}

This appendix provides a non-parametric sanity check on the
Stage-3 feature ordering reported in
Section~\ref{sec:disparity:variance}. It is not intended as a
standalone imputation method or as an independent estimator: the tree
is trained on the same per-cell frame and the same feature set the
Bayesian model uses, and is reported only to confirm that the
hierarchy of feature contributions is recoverable from a model with
very different inductive biases.

\paragraph{Setup.}
We fit a gradient-boosted regressor (\texttt{XGBoost}, $600$ trees,
depth $4$, learning rate $0.05$) on the tidy frame of
$(\text{model}, \text{benchmark}, \text{language})$ cells.
Continuous features enter standardised; model, language, task,
script, and language family enter as one-hot encodings. For each of
$5$ random seeds and each (benchmark, resource tier) combination with
$\geq 2$ languages, one language is held out and its seven model
rows form the test set; the held-out language never appears for that
benchmark in training.

\paragraph{Held-out error and feature ranking.}
Table~\ref{tab:tree-holdout} reports per-benchmark MAE/RMSE.
Pooled across held-out cells the imputation lies within
$3.98 \pm 0.83$ points (NLU) and $6.92 \pm 1.34$ points (Reasoning)
of the true accuracy on the $[0,1]$ scale; the per-tier breakdown
(Table~\ref{tab:tree-holdout-joshi}) shows graceful degradation
from Class~4 to Class~1 rather than collapse on low-resource tiers.
SHAP-based feature importance
(Table~\ref{tab:tree-importance}, categorical one-hots summed back
to their parent feature) places \textsc{repr\_sim\_en} as the
largest linguistic contributor in both buckets ($19.9\%$ NLU,
$29.5\%$ Reasoning), followed by fertility and syntactic distance,
with phonological distance and \textsc{resource} class trailing.
This ranking matches the Stage-3 credible-interval ordering in
Section~\ref{sec:disparity:variance}.

\paragraph{Caveats.}
Feature selection is shared with the Bayesian model, so the
agreement should be read as internal consistency rather than
independent corroboration. The tree is evaluated on held-out
languages from \emph{seen} benchmarks; generalisation to entirely
unseen benchmarks or model families is not tested here. We report
no baselines (e.g.\ language-mean or \textsc{resource tier}-only
regressors); the held-out numbers should accordingly be read as
an upper bound on what the feature set affords, not as a benchmarked
imputation result.

\begin{table}[!htbp]
\centering
\small
\setlength{\tabcolsep}{5pt}
\renewcommand{\arraystretch}{1.15}
\resizebox{\columnwidth}{!}{%
\begin{tabular}{lcc}
\toprule
\textbf{Benchmark} & \textbf{MAE} & \textbf{RMSE} \\
\midrule
\multicolumn{3}{l}{\cellcolor{bucketNLU!55}\textbf{NLU}} \\
\midrule
\textsc{okapi-mmlu}         & $0.019 \pm 0.003$ & $0.026 \pm 0.008$ \\
\textsc{mmlu-indic-roman}   & $0.021 \pm 0.008$ & $0.025 \pm 0.008$ \\
\textsc{milu}               & $0.022 \pm 0.006$ & $0.025 \pm 0.006$ \\
\textsc{boolq-indic}        & $0.030 \pm 0.010$ & $0.040 \pm 0.012$ \\
\textsc{mmmlu}              & $0.031 \pm 0.011$ & $0.038 \pm 0.013$ \\
\textsc{global-mmlu}        & $0.039 \pm 0.019$ & $0.050 \pm 0.025$ \\
\textsc{triviaqa-indic-mcq} & $0.047 \pm 0.036$ & $0.058 \pm 0.041$ \\
\textsc{include}            & $0.053 \pm 0.024$ & $0.065 \pm 0.030$ \\
\textsc{mmlu-prox}          & $0.097 \pm 0.066$ & $0.127 \pm 0.073$ \\
\midrule
\rowcolor{bucketNLU!55} \textbf{NLU overall} & $\mathbf{0.040 \pm 0.008}$ & $\mathbf{0.063 \pm 0.020}$ \\
\midrule
\multicolumn{3}{l}{\cellcolor{bucketReason!55}\textbf{Reasoning}} \\
\midrule
\textsc{xcopa}       & $0.034 \pm 0.027$ & $0.041 \pm 0.033$ \\
\textsc{xstorycloze} & $0.063 \pm 0.024$ & $0.076 \pm 0.028$ \\
\textsc{belebele}    & $0.065 \pm 0.027$ & $0.078 \pm 0.033$ \\
\textsc{xwinograd}   & $0.075 \pm 0.021$ & $0.083 \pm 0.021$ \\
\textsc{gsm8k-indic} & $0.084 \pm 0.038$ & $0.102 \pm 0.045$ \\
\textsc{mgsm}        & $0.094 \pm 0.054$ & $0.115 \pm 0.052$ \\
\midrule
\rowcolor{bucketReason!55} \textbf{Reasoning overall} & $\mathbf{0.069 \pm 0.013}$ & $\mathbf{0.091 \pm 0.017}$ \\
\bottomrule
\end{tabular}%
}
\caption{Per-benchmark held-out imputation error. For each of $5$
seeds, one language per benchmark is held out (all $7$ model rows for
that language form the test set); MAE/RMSE on the $[0,1]$ scale,
mean$\pm$std across seeds. Bucket shading as in main-text tables
(\colorbox{bucketNLU}{NLU}, \colorbox{bucketReason}{Reasoning}).}
\label{tab:tree-holdout}
\end{table}
\begin{table}[!htbp]
\centering
\small
\setlength{\tabcolsep}{5pt}
\renewcommand{\arraystretch}{1.15}
\begin{tabular}{lccc}
\toprule
\textbf{Resource tier} & \textbf{$n$} & \textbf{MAE} & \textbf{RMSE} \\
\midrule
\multicolumn{4}{l}{\cellcolor{bucketNLU!55}\textbf{NLU}} \\
\midrule
1-Scraping  &  5 & $0.070 \pm 0.042$ & $0.083 \pm 0.048$ \\
2-Hopefuls  & 45 & $0.040 \pm 0.023$ & $0.050 \pm 0.028$ \\
3-Rising    & 25 & $0.044 \pm 0.031$ & $0.054 \pm 0.035$ \\
4-Underdogs & 25 & $0.038 \pm 0.029$ & $0.048 \pm 0.038$ \\
\midrule
\multicolumn{4}{l}{\cellcolor{bucketReason!55}\textbf{Reasoning}} \\
\midrule
1-Scraping  &  5 & $0.106 \pm 0.049$ & $0.123 \pm 0.053$ \\
2-Hopefuls  & 25 & $0.097 \pm 0.034$ & $0.115 \pm 0.038$ \\
3-Rising    & 15 & $0.051 \pm 0.024$ & $0.062 \pm 0.025$ \\
4-Underdogs & 25 & $0.050 \pm 0.028$ & $0.059 \pm 0.029$ \\
\bottomrule
\end{tabular}
\caption{Held-out imputation error stratified by the held-out
language's resource tier; $n$ is the number of held-out
(language, benchmark) cells across $5$ seeds. Class~$5$ is omitted
(English is the only Class~$5$ language and has no within-tier
holdout partner). Shading as in main-text tables
(\colorbox{bucketNLU}{NLU}, \colorbox{bucketReason}{Reasoning}).}
\label{tab:tree-holdout-joshi}
\end{table}

\begin{table}[!htbp]
\centering
\small
\setlength{\tabcolsep}{5pt}
\renewcommand{\arraystretch}{1.15}
\begin{tabular}{lcc}
\toprule
\textbf{Feature group} &
\cellcolor{bucketNLU!55}\textbf{NLU} &
\cellcolor{bucketReason!55}\textbf{Reasoning} \\
\midrule
\multicolumn{3}{l}{\textit{Linguistic / model--language features}} \\
$\textsc{repr\_sim\_en}$ (CKA) & \textbf{19.9\%} & \textbf{29.5\%} \\
$\textsc{fertility}$           & 6.5\%           & 4.5\%           \\
$d^{\text{syn}}_{\text{en}}$   & 4.6\%           & 5.9\%           \\
$\textsc{joshi\_class}$        & 3.4\%           & 2.6\%           \\
$d^{\text{phon}}_{\text{en}}$  & 0.5\%           & 0.8\%           \\
\midrule
\multicolumn{3}{l}{\textit{Language identity (categorical)}} \\
$\text{language}$              & 4.3\%           & 5.1\%           \\
$\text{family}$                & 2.3\%           & 3.8\%           \\
$\text{script}$                & 1.8\%           & 1.7\%           \\
\midrule
\multicolumn{3}{l}{\textit{Model- and task-level identifiers}} \\
$\text{model}$                 & 31.7\%          & 28.5\%          \\
$\text{task}$                  & 24.9\%          & 17.6\%          \\
\bottomrule
\end{tabular}
\caption{Tree feature importance: share of total SHAP magnitude per
feature group (one-hot encodings summed back to parents). Shading as
in main-text variance decomposition
(\colorbox{bucketNLU}{NLU}, \colorbox{bucketReason}{Reasoning});
\textsc{repr\_sim\_en} is the largest linguistic contributor in both
buckets, reproducing the Stage-3 Bayesian ordering
(\S\ref{sec:disparity:variance}).}
\label{tab:tree-importance}
\end{table}

\section{NLG bucket --- why it is excluded from the main text}
\label{app:nlg}

The NLG bucket is the open-generation analogue of the analysis in the
main text: $413$ rows of chrF scores from FLORES and IN22-conv
translation and IGB-XSum summarisation, on $12$ Indic and Dravidian
languages, with the same seven models. The fit produces a credibly
positive coefficient on \texttt{repr\_sim\_en\_z}
($\beta = +0.053$ $[+0.025, +0.080]$), consistent with NLU and
Reasoning, but the headline disparity-explanation metric
$R^2_{\text{ling}}$ is not interpretable on this bucket.

\paragraph{The identifiability failure.}
Adding the structured language features inflates rather than shrinks
$\sigma_\alpha$ ($0.014 \to 0.304$, a $22\times$ expansion), which
produces a strongly negative $R^2_{\text{ling}}$. The mechanism is
categorical sparsity: of the $12$ languages in the pool, $9$ of $10$
scripts are singletons (Bengali, Gujarati, Kannada, Malayalam, Odia,
Gurmukhi, Tamil, Telugu, Arabic) and the only two language families
present are Indic ($8$) and Dravidian ($4$). With this structure, the
$C(\mathrm{script})$ and $C(\mathrm{family})$ categorical effects are
collinear with the language random intercept $\alpha_\ell$ for those
languages, and the model cannot identify them separately.

\paragraph{Diagnostic implication.}
The variance share of the language residual is $55.5\%$ in NLG
 but this is an artefact of the same
identifiability failure, not a substantive finding about Indic and
Dravidian languages. Reading the NLG variance decomposition as
``language matters most'' would be a misinterpretation.

\paragraph{What would unblock NLG.}
The fix is not methodological: it is data coverage. Adding a single
non-Indic, non-Dravidian generation benchmark (e.g.\ FLORES
\texttt{en$\leftrightarrow$es}, \texttt{en$\leftrightarrow$zh}, or
\texttt{en$\leftrightarrow$ar}) would bring the script-family
categoricals out of singleton status and restore $R^2_{\text{ling}}$
identifiability. Until that is done, the NLG analysis should be read
as ``\texttt{repr\_sim\_en} predicts generation quality on Indic
languages'' rather than as a parallel disparity decomposition.

\begin{table}[!h]
\centering
\footnotesize
\setlength{\tabcolsep}{4pt}
\renewcommand{\arraystretch}{1.15}
\newcommand{\bk}[1]{\textsc{#1}}

\begin{tabularx}{\columnwidth}{@{}
  >{\raggedright\arraybackslash}p{0.18\columnwidth}
  >{\raggedright\arraybackslash}X
  >{\raggedright\arraybackslash}p{0.20\columnwidth}@{}}
\toprule
\textbf{Bucket} & \textbf{Benchmarks} & \textbf{Score} \\
\midrule
\rowcolor{bucketNLG!55}
\textbf{NLG} \newline \textit{(open gen.)} &
\bk{igb-flores} (FLORES-200 translation, both dir.),
\bk{in22-conv-16k} (IN22 conv.\ translation, both dir.),
\bk{igb-xsum} (cross-lingual summarisation) &
chrF~$\in[0,100]$, rescaled to $[0,1]$ \\
\bottomrule
\end{tabularx}

\caption{NLG benchmarks (\colorbox{bucketNLG}{NLG}); companion to
Table~\ref{tab:buckets}. Translation source$\rightarrow$target
directions are preserved as separate modeling tasks so that
$\textsc{task:language}$ interactions are identifiable.}
\label{tab:buckets_nlg}
\end{table}

\begin{table}[!h]
\centering
\small
\setlength{\tabcolsep}{5pt}
\renewcommand{\arraystretch}{1.15}

\resizebox{\columnwidth}{!}{%
\begin{tabular}{lrrrrrl}
\toprule
\textbf{Benchmark} & $k$ (lang) & $m$ & $\chi^2$ & $p$ & Kendall $W$ & sig \\
\midrule
\rowcolor{bucketNLG!55} \textsc{igb-flores}    & 29 & 7 & 148.4 & $1.7\!\times\!10^{-29}$ & 0.757 & *** \\
\rowcolor{bucketNLG!55} \textsc{igb-xsum}      & 28 & 7 & 154.4 & $9.3\!\times\!10^{-31}$ & 0.817 & *** \\
\rowcolor{bucketNLG!55} \textsc{in22-conv-16k} & 22 & 7 & 107.2 & $7.8\!\times\!10^{-21}$ & 0.729 & *** \\
\bottomrule
\end{tabular}%
}

\caption{Per-benchmark Friedman test on language scores with models as judges
(rows = languages, columns = models, balanced via dropna) for the NLG bucket.
Kendall's $W = \chi^2 / (m(k-1))$ is the $[0,1]$ effect-size analog.
Companion to Table~\ref{tab:friedman_per_benchmark}.}
\label{tab:friedman_per_benchmark_nlg}
\end{table}

% \begin{table}[!h]
% \centering
% \small
% \setlength{\tabcolsep}{5pt}
% \renewcommand{\arraystretch}{1.15}
% \begin{tabular}{lrrrrrl}
% \toprule
% \textbf{Benchmark} & $k$ (lang) & $m$ & $\chi^2$ & $p$ & Kendall $W$ & sig \\
% \midrule
% \rowcolor{bucketNLG!55} \textsc{igb-flores}    & 29 & 7 & 148.4 & $1.7\!\times\!10^{-29}$ & 0.757 & *** \\
% \rowcolor{bucketNLG!55} \textsc{igb-xsum}      & 28 & 7 & 154.4 & $9.3\!\times\!10^{-31}$ & 0.817 & *** \\
% \rowcolor{bucketNLG!55} \textsc{in22-conv-16k} & 22 & 7 & 107.2 & $7.8\!\times\!10^{-21}$ & 0.729 & *** \\
% \bottomrule
% \end{tabular}
% \caption{Per-benchmark Friedman test on language scores with models as judges (rows = languages, columns = models, balanced via dropna) for the NLG bucket. Kendall's $W = \chi^2 / (m\,(k-1))$ is the $[0,1]$
% effect-size analog. Companion to
% Table~\ref{tab:friedman_per_benchmark}.}
% \label{tab:friedman_per_benchmark_nlg}
% \end{table}

\section{Full Dunn's pairwise post-hoc matrices}
\label{app:dunn_full}

Each table reports Dunn's pairwise post-hoc test for one (model, bucket) combination, comparing the
five resource tiers \cite{joshi-etal-2020-state}:
1=Scraping, 2=Hopefuls, 3=Rising, 4=Underdogs, 5=Winners. Entries
are Benjamini--Hochberg FDR-adjusted $p$-values for the null hypothesis that two tiers have
the same distribution of within-$(\text{benchmark},
\text{task\_id})$ $z$-normalised per-language scores. The matrices
are symmetric and the diagonal is omitted. Significance markers
follow the usual convention: \textit{***}~$p<.001$,
\textit{**}~$p<.01$, \textit{*}~$p<.05$, and \textit{ns}~for
$p\geq.05$; $p$-values below $10^{-4}$ are written as
``$<\!10^{-4}$''. A significant off-diagonal cell means the two
tiers differ after correcting for all ten pairwise comparisons; an
\textit{ns} cell means the data do not separate them. Read together
with the omnibus Kruskal--Wallis result in
Table~\ref{tab:kw_joshi_tiers}, these per-model matrices show
\emph{where} on the resource ladder each model's tier ordering is
actually resolved versus where adjacent tiers collapse.

% ============================================================
%  Dunn's post-hoc on Joshi resource classes
%  BH-corrected pairwise p-values. *** p<.001, ** p<.01, * p<.05, ns p>=.05.
%  Class labels (Joshi et al., 2020): 1=Scraping, 2=Hopefuls, 3=Rising,
%  4=Underdogs, 5=Winners.  All p-values are symmetric; matrices show full grid
%  for readability.  Source: scikit_posthocs.posthoc_dunn(..., p_adjust="fdr_bh")
%  on within-(model, benchmark) z-scored per-language scores.
% ============================================================

% Reusable compact column type
\newcommand{\dcell}[1]{\multicolumn{1}{c}{#1}}

\begin{table}[!h]
\centering
\scriptsize
\setlength{\tabcolsep}{3pt}
\begin{tabular}{l ccccc}
\toprule
 & 1 & 2 & 3 & 4 & 5 \\
\midrule
1 & ---                 & $<\!10^{-4}$\,***  & $<\!10^{-4}$\,***  & $<\!10^{-4}$\,***  & $<\!10^{-4}$\,*** \\
2 & $<\!10^{-4}$\,***   & ---                & $<\!10^{-4}$\,***  & $<\!10^{-4}$\,***  & $<\!10^{-4}$\,*** \\
3 & $<\!10^{-4}$\,***   & $<\!10^{-4}$\,***  & ---                & .0886\,ns          & $<\!10^{-4}$\,*** \\
4 & $<\!10^{-4}$\,***   & $<\!10^{-4}$\,***  & .0886\,ns          & ---                & .0001\,*** \\
5 & $<\!10^{-4}$\,***   & $<\!10^{-4}$\,***  & $<\!10^{-4}$\,***  & .0001\,***         & --- \\
\bottomrule
\end{tabular}
\caption{Qwen3.5-122B-A10B, NLU.}
\label{tab:dunn_qwen122b_nlu}
\end{table}

\begin{table}[!ht]
\centering
\scriptsize
\setlength{\tabcolsep}{3pt}
\begin{tabular}{l ccccc}
\toprule
 & 1 & 2 & 3 & 4 & 5 \\
\midrule
1 & ---           & .0311\,*           & $<\!10^{-4}$\,***  & $<\!10^{-4}$\,***  & $<\!10^{-4}$\,*** \\
2 & .0311\,*      & ---                & $<\!10^{-4}$\,***  & $<\!10^{-4}$\,***  & $<\!10^{-4}$\,*** \\
3 & $<\!10^{-4}$\,*** & $<\!10^{-4}$\,*** & ---            & $<\!10^{-4}$\,***  & $<\!10^{-4}$\,*** \\
4 & $<\!10^{-4}$\,*** & $<\!10^{-4}$\,*** & $<\!10^{-4}$\,*** & ---             & .0021\,** \\
5 & $<\!10^{-4}$\,*** & $<\!10^{-4}$\,*** & $<\!10^{-4}$\,*** & .0021\,**       & --- \\
\bottomrule
\end{tabular}
\caption{Qwen3.5-4B, NLU.}
\label{tab:dunn_qwen4b_nlu}
\end{table}

\begin{table}[!ht]
\centering
\scriptsize
\setlength{\tabcolsep}{3pt}
\begin{tabular}{l ccccc}
\toprule
 & 1 & 2 & 3 & 4 & 5 \\
\midrule
1 & ---                 & .0003\,***         & $<\!10^{-4}$\,***  & $<\!10^{-4}$\,***  & $<\!10^{-4}$\,*** \\
2 & .0003\,***          & ---                & $<\!10^{-4}$\,***  & $<\!10^{-4}$\,***  & $<\!10^{-4}$\,*** \\
3 & $<\!10^{-4}$\,***   & $<\!10^{-4}$\,***  & ---                & $<\!10^{-4}$\,***  & $<\!10^{-4}$\,*** \\
4 & $<\!10^{-4}$\,***   & $<\!10^{-4}$\,***  & $<\!10^{-4}$\,***  & ---                & .0328\,* \\
5 & $<\!10^{-4}$\,***   & $<\!10^{-4}$\,***  & $<\!10^{-4}$\,***  & .0328\,*           & --- \\
\bottomrule
\end{tabular}
\caption{aya-expanse-32b, NLU.}
\label{tab:dunn_aya32b_nlu}
\end{table}

\begin{table}[!ht]
\centering
\scriptsize
\setlength{\tabcolsep}{3pt}
\begin{tabular}{l ccccc}
\toprule
 & 1 & 2 & 3 & 4 & 5 \\
\midrule
1 & ---                & $<\!10^{-4}$\,***  & $<\!10^{-4}$\,***  & $<\!10^{-4}$\,***  & $<\!10^{-4}$\,*** \\
2 & $<\!10^{-4}$\,***  & ---                & $<\!10^{-4}$\,***  & $<\!10^{-4}$\,***  & $<\!10^{-4}$\,*** \\
3 & $<\!10^{-4}$\,***  & $<\!10^{-4}$\,***  & ---                & $<\!10^{-4}$\,***  & $<\!10^{-4}$\,*** \\
4 & $<\!10^{-4}$\,***  & $<\!10^{-4}$\,***  & $<\!10^{-4}$\,***  & ---                & .0011\,** \\
5 & $<\!10^{-4}$\,***  & $<\!10^{-4}$\,***  & $<\!10^{-4}$\,***  & .0011\,**          & --- \\
\bottomrule
\end{tabular}
\caption{gpt-oss-20b, NLU.}
\label{tab:dunn_gptoss_nlu}
\end{table}

\begin{table}[!ht]
\centering
\scriptsize
\setlength{\tabcolsep}{3pt}
\begin{tabular}{l ccccc}
\toprule
 & 1 & 2 & 3 & 4 & 5 \\
\midrule
1 & ---                & $<\!10^{-4}$\,***  & $<\!10^{-4}$\,***  & $<\!10^{-4}$\,***  & $<\!10^{-4}$\,*** \\
2 & $<\!10^{-4}$\,***  & ---                & .5950\,ns          & $<\!10^{-4}$\,***  & $<\!10^{-4}$\,*** \\
3 & $<\!10^{-4}$\,***  & .5950\,ns          & ---                & $<\!10^{-4}$\,***  & $<\!10^{-4}$\,*** \\
4 & $<\!10^{-4}$\,***  & $<\!10^{-4}$\,***  & $<\!10^{-4}$\,***  & ---                & $<\!10^{-4}$\,*** \\
5 & $<\!10^{-4}$\,***  & $<\!10^{-4}$\,***  & $<\!10^{-4}$\,***  & $<\!10^{-4}$\,***  & --- \\
\bottomrule
\end{tabular}
\caption{sarvam-105b, NLU.}
\label{tab:dunn_sarvam105_nlu}
\end{table}

\begin{table}[!ht]
\centering
\scriptsize
\setlength{\tabcolsep}{3pt}
\begin{tabular}{l ccccc}
\toprule
 & 1 & 2 & 3 & 4 & 5 \\
\midrule
1 & ---                & $<\!10^{-4}$\,***  & $<\!10^{-4}$\,***  & $<\!10^{-4}$\,***  & $<\!10^{-4}$\,*** \\
2 & $<\!10^{-4}$\,***  & ---                & .3295\,ns          & .4224\,ns          & $<\!10^{-4}$\,*** \\
3 & $<\!10^{-4}$\,***  & .3295\,ns          & ---                & .6598\,ns          & $<\!10^{-4}$\,*** \\
4 & $<\!10^{-4}$\,***  & .4224\,ns          & .6598\,ns          & ---                & $<\!10^{-4}$\,*** \\
5 & $<\!10^{-4}$\,***  & $<\!10^{-4}$\,***  & $<\!10^{-4}$\,***  & $<\!10^{-4}$\,***  & --- \\
\bottomrule
\end{tabular}
\caption{sarvam-30b, NLU.}
\label{tab:dunn_sarvam30_nlu}
\end{table}

\begin{table}[!ht]
\centering
\scriptsize
\setlength{\tabcolsep}{3pt}
\begin{tabular}{l ccccc}
\toprule
 & 1 & 2 & 3 & 4 & 5 \\
\midrule
1 & ---                & .0862\,ns          & $<\!10^{-4}$\,***  & $<\!10^{-4}$\,***  & $<\!10^{-4}$\,*** \\
2 & .0862\,ns          & ---                & $<\!10^{-4}$\,***  & $<\!10^{-4}$\,***  & $<\!10^{-4}$\,*** \\
3 & $<\!10^{-4}$\,***  & $<\!10^{-4}$\,***  & ---                & $<\!10^{-4}$\,***  & .0001\,*** \\
4 & $<\!10^{-4}$\,***  & $<\!10^{-4}$\,***  & $<\!10^{-4}$\,***  & ---                & .0830\,ns \\
5 & $<\!10^{-4}$\,***  & $<\!10^{-4}$\,***  & .0001\,***         & .0830\,ns          & --- \\
\bottomrule
\end{tabular}
\caption{tiny-aya-global, NLU.}
\label{tab:dunn_tinyaya_nlu}
\end{table}

% ---------- PER-MODEL REASONING ----------

\begin{table}[!ht]
\centering
\scriptsize
\setlength{\tabcolsep}{3pt}
\begin{tabular}{l ccccc}
\toprule
 & 1 & 2 & 3 & 4 & 5 \\
\midrule
1 & ---         & .8552\,ns          & .0142\,*           & .0105\,*           & .0260\,* \\
2 & .8552\,ns   & ---                & $<\!10^{-4}$\,***  & $<\!10^{-4}$\,***  & .0105\,* \\
3 & .0142\,*    & $<\!10^{-4}$\,***  & ---                & .8007\,ns          & .8007\,ns \\
4 & .0105\,*    & $<\!10^{-4}$\,***  & .8007\,ns          & ---                & .8007\,ns \\
5 & .0260\,*    & .0105\,*           & .8007\,ns          & .8007\,ns          & --- \\
\bottomrule
\end{tabular}
\caption{Qwen3.5-122B-A10B, Reasoning.}
\label{tab:dunn_qwen122b_reasoning}
\end{table}

\begin{table}[!ht]
\centering
\scriptsize
\setlength{\tabcolsep}{3pt}
\begin{tabular}{l ccccc}
\toprule
 & 1 & 2 & 3 & 4 & 5 \\
\midrule
1 & ---           & .9191\,ns          & .0210\,*           & .0010\,***         & .0010\,*** \\
2 & .9191\,ns     & ---                & $<\!10^{-4}$\,***  & $<\!10^{-4}$\,***  & $<\!10^{-4}$\,*** \\
3 & .0210\,*      & $<\!10^{-4}$\,***  & ---                & .0408\,*           & .0394\,* \\
4 & .0010\,***    & $<\!10^{-4}$\,***  & .0408\,*           & ---                & .2596\,ns \\
5 & .0010\,***    & $<\!10^{-4}$\,***  & .0394\,*           & .2596\,ns          & --- \\
\bottomrule
\end{tabular}
\caption{Qwen3.5-4B, Reasoning.}
\label{tab:dunn_qwen4b_reasoning}
\end{table}

\begin{table}[!ht]
\centering
\scriptsize
\setlength{\tabcolsep}{3pt}
\begin{tabular}{l ccccc}
\toprule
 & 1 & 2 & 3 & 4 & 5 \\
\midrule
1 & ---           & .5501\,ns          & .0464\,*           & .0015\,**          & .0002\,*** \\
2 & .5501\,ns     & ---                & .0088\,**          & $<\!10^{-4}$\,***  & $<\!10^{-4}$\,*** \\
3 & .0464\,*      & .0088\,**          & ---                & .0071\,**          & .0016\,** \\
4 & .0015\,**     & $<\!10^{-4}$\,***  & .0071\,**          & ---                & .0500\,ns \\
5 & .0002\,***    & $<\!10^{-4}$\,***  & .0016\,**          & .0500\,ns          & --- \\
\bottomrule
\end{tabular}
\caption{aya-expanse-32b, Reasoning.}
\label{tab:dunn_aya32b_reasoning}
\end{table}

\begin{table}[!ht]
\centering
\scriptsize
\setlength{\tabcolsep}{3pt}
\begin{tabular}{l ccccc}
\toprule
 & 1 & 2 & 3 & 4 & 5 \\
\midrule
1 & ---         & .0755\,ns   & .9072\,ns   & .2782\,ns   & .0240\,* \\
2 & .0755\,ns   & ---         & .0004\,***  & .1282\,ns   & .2250\,ns \\
3 & .9072\,ns   & .0004\,***  & ---         & .0240\,*    & .0018\,** \\
4 & .2782\,ns   & .1282\,ns   & .0240\,*    & ---         & .0431\,* \\
5 & .0240\,*    & .2250\,ns   & .0018\,**   & .0431\,*    & --- \\
\bottomrule
\end{tabular}
\caption{gpt-oss-20b, Reasoning.}
\label{tab:dunn_gptoss_reasoning}
\end{table}

\begin{table}[!ht]
\centering
\scriptsize
\setlength{\tabcolsep}{3pt}
\begin{tabular}{l ccccc}
\toprule
 & 1 & 2 & 3 & 4 & 5 \\
\midrule
1 & ---         & .1318\,ns   & .8575\,ns   & .1318\,ns   & .0268\,* \\
2 & .1318\,ns   & ---         & .0073\,**   & .8575\,ns   & .1562\,ns \\
3 & .8575\,ns   & .0073\,**   & ---         & .0063\,**   & .0063\,** \\
4 & .1318\,ns   & .8575\,ns   & .0063\,**   & ---         & .1318\,ns \\
5 & .0268\,*    & .1562\,ns   & .0063\,**   & .1318\,ns   & --- \\
\bottomrule
\end{tabular}
\caption{sarvam-105b, Reasoning.}
\label{tab:dunn_sarvam105_reasoning}
\end{table}

\begin{table}[t]
\centering
\scriptsize
\setlength{\tabcolsep}{3pt}
\begin{tabular}{l ccccc}
\toprule
 & 1 & 2 & 3 & 4 & 5 \\
\midrule
1 & ---         & .0349\,*   & .4451\,ns  & .1495\,ns  & .1325\,ns \\
2 & .0349\,*    & ---        & .0027\,**  & .1053\,ns  & .8418\,ns \\
3 & .4451\,ns   & .0027\,**  & ---        & .1495\,ns  & .1495\,ns \\
4 & .1495\,ns   & .1053\,ns  & .1495\,ns  & ---        & .4253\,ns \\
5 & .1325\,ns   & .8418\,ns  & .1495\,ns  & .4253\,ns  & --- \\
\bottomrule
\end{tabular}
\caption{sarvam-30b, Reasoning.}
\label{tab:dunn_sarvam30_reasoning}
\end{table}

\begin{table}[!ht]
\centering
\scriptsize
\setlength{\tabcolsep}{3pt}
\begin{tabular}{l ccccc}
\toprule
 & 1 & 2 & 3 & 4 & 5 \\
\midrule
1 & ---           & .6452\,ns          & .0209\,*           & .0015\,**          & .0002\,*** \\
2 & .6452\,ns     & ---                & .0006\,***         & $<\!10^{-4}$\,***  & $<\!10^{-4}$\,*** \\
3 & .0209\,*      & .0006\,***         & ---                & .0531\,ns          & .0044\,** \\
4 & .0015\,**     & $<\!10^{-4}$\,***  & .0531\,ns          & ---                & .0488\,* \\
5 & .0002\,***    & $<\!10^{-4}$\,***  & .0044\,**          & .0488\,*           & --- \\
\bottomrule
\end{tabular}
\caption{tiny-aya-global, Reasoning.}
\label{tab:dunn_tinyaya_reasoning}
\end{table}

\clearpage

\section{Compute}
All large language model evaluations were executed using the 
\textsc{lm-evaluation-harness} framework across a single computational 
node equipped with eight NVIDIA B200 GPUs. Downstream statistical 
estimation and posterior inference for the nested Bayesian hierarchical 
framework were performed subsequently, with the sampling process requiring 
approximately 1.5 hours of compute time per modelling stage.

\end{document}